%% file: main_icra.tex
\documentclass[letterpaper, 10 pt, conference]{ieeeconf}  
                                                          
\include{preamble}
\usepackage{caption}
\usepackage{wrapfig}

\newcommand{\TODO}[1]{\textbf{\textcolor{red}{[TODO: #1]}}}

\newcommand{\SB}[1]{\textbf{\textcolor{cyan}{[SB: #1]}}}
\newcommand{\DS}[1]{\textbf{\textcolor{green}{[DS: #1]}}}
\newcommand{\TB}[1]{\textbf{\textcolor{blue}{[TB: #1]}}}
\newcommand{\EG}[1]{\textbf{\textcolor{purple}{[EG: #1]}}}

\newcommand{\SK}[1]{\textbf{\textcolor{violet}{[TC: #1]}}}

\newcommand{\REV}[1]{\textcolor{blue}{#1}}

\renewcommand{\TODO}[1]{}
\renewcommand{\SB}[1]{}
\renewcommand{\DS}[1]{}
\renewcommand{\TB}[1]{}
\renewcommand{\EG}[1]{}
\renewcommand{\SK}[1]{}

\renewcommand{\REV}[1]{#1}

\renewcommand{\baselinestretch}{0.98}
\begin{document}

\title{\LARGE \bf
Balancing Efficiency and Comfort in Robot-Assisted Bite Transfer
}

\author{Suneel Belkhale$^{1}$, Ethan K. Gordon$^{2}$, Yuxiao Chen$^{1}$, \\Siddhartha Srinivasa$^{2}$, Tapomayukh Bhattacharjee$^{3}$, Dorsa Sadigh$^{1}$\\ \text{\footnotesize Stanford University$^{1}$,\, University of Washington$^{2}$,\, Cornell University$^{3}$}}




%

\twocolumn[{%
\renewcommand\twocolumn[1][]{#1}%
\maketitle

\vspace{-15pt}

\begin{center}
    \includegraphics[width=0.92\textwidth, trim={10pt 5pt 5pt 0pt}, clip]{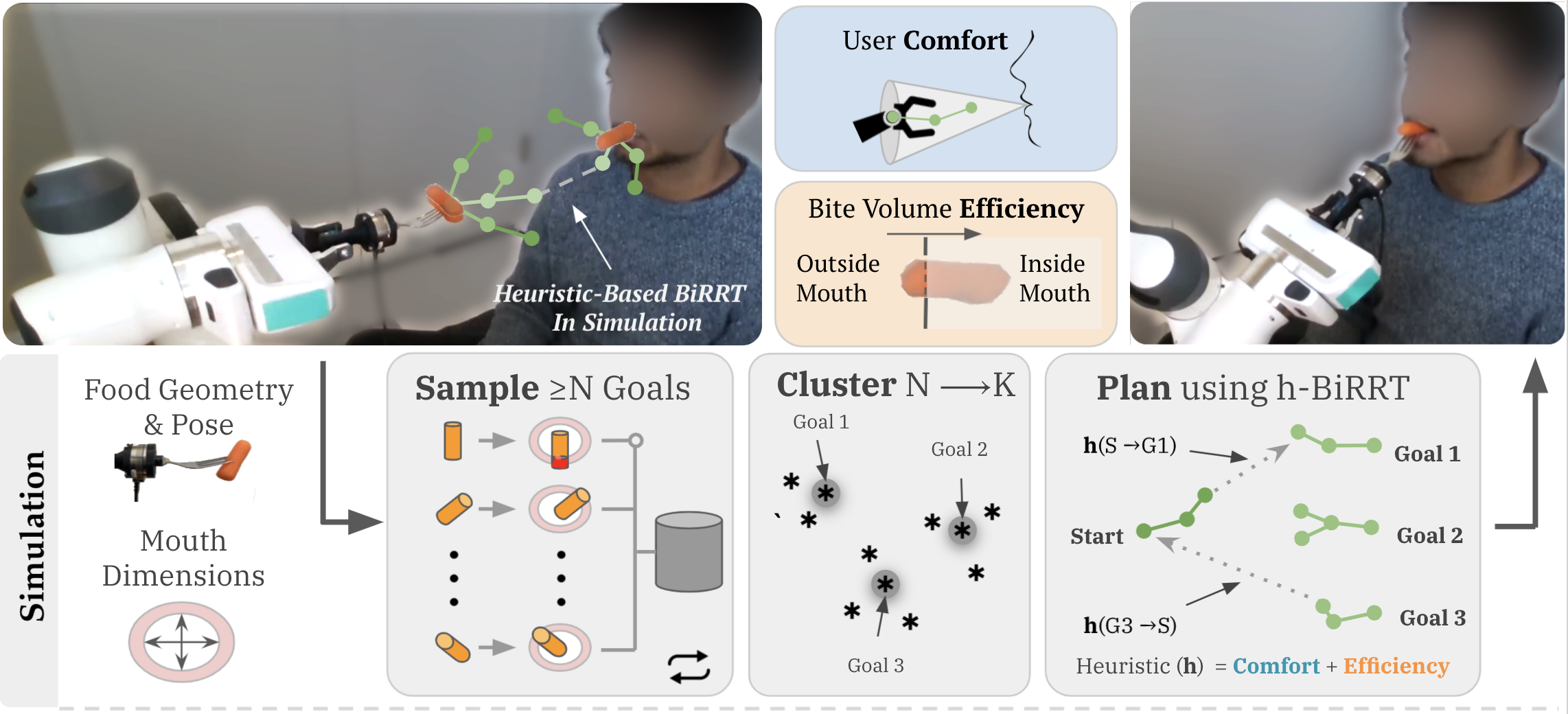}
    \captionof{figure}{Our method finds feasible bite transfer trajectories in simulation. Given the food geometry and pose on the fork, we sample at least $N$ goal food poses that are checked for collisions with the mouth geometry using a learned constraint model. Next, we cluster the goal poses and use heuristic-guided BiRRT to reach cluster centroids with comfort (blue) and bite volume efficiency (orange) heuristics. 
    }
    \label{fig:front}
\end{center}
}]




\begin{abstract}
Robot-assisted feeding in household environments is challenging because it requires robots to generate trajectories that effectively bring food items of varying shapes and sizes into the mouth while making sure the user is comfortable. Our key insight is that in order to solve this challenge, robots must balance the \emph{efficiency} of feeding a food item with the \emph{comfort} of each individual bite. We formalize comfort and efficiency as heuristics to incorporate in motion planning. We present an approach based on heuristics-guided bi-directional Rapidly-exploring Random Trees (h-BiRRT) that selects bite transfer trajectories of arbitrary food item geometries and shapes using our developed bite efficiency and comfort heuristics and a learned constraint model. Real-robot evaluations show that optimizing both comfort and efficiency significantly outperforms a fixed-pose based method, and users preferred our method significantly more than that of a method that maximizes only user comfort. Videos and Appendices are found on our website: \url{https://tinyurl.com/bticra22}.
\end{abstract}

\section{Introduction}
\label{sect:intro}


Imagine a setting where you want to pick up a piece of food, e.g., a baby carrot, from a salad bowl to eat. \REV{Non-disabled people might overlook the complexity of this daily task} --- they might use a fork to pick up the carrot, while carrying a conversation and not paying as much attention on how the carrot is placed on the fork. \REV{Regardless of this placement, they move the fork in a manner that is not only efficient in how much food can be eaten but is also comfortable for the duration of the motion. This task presents numerous challenges for more than 12 million people with mobility-related disabilities \cite{brault2012americans}.} Assistive robot arms have the potential to bridge this gap, and therefore provide care for those with disabilities.
However, operating these arms can be challenging \cite{losey2020controlling, li2020multi}. In our initial surveys, \REV{people with mobility impairment} mentioned the need for intelligent autonomy that optimizes comfort and adapts to the food item being fed. We envision intelligent algorithms that are aware of user comfort without the need for explicit user input. Achieving this level of autonomy presents a number of challenges which carry over to other robotics applications, including: 
1) \textit{perceiving} and choosing the next bite of food on a plate, 2) \textit{acquiring} the food item with an appropriate tool, 3) \textit{transferring} these items into the mouth in an efficient and comfortable manner. In recent years, there has been significant advances in food perception and acquisition \cite{2019Feng, 2019Gallenberger}. It turns out that the food acquisition strategy (e.g. fork skewering angle) heavily affects a user's comfort during bite transfer~\cite{2019Gallenberger}; however, prior bite transfer methods rely on predetermined transfer trajectories for a discrete set of acquisition strategies and food geometries~\cite{bhattacharjee2018food}.


To handle a wide variety of food items and acquisition methods, a bite transfer strategy must optimize its trajectories on the fly by bringing food into a mouth without sacrificing user comfort. However, this is challenging with real world sources of variation (e.g. food geometries, sizes, acquisition poses on the fork, and mouth shapes). Even with one food geometry and acquisition pose, there are often many different ``collision-free" paths into the mouth, so the feeding agent should filter this solution space intelligently. 
For instance, consider a vertically aligned baby carrot oriented perpendicular to the fork, as shown in Fig.~\ref{fig:front}.
There are a wide range of possible feeding paths; some may come too close to a person's face, affecting comfort, while others may only bring the tip of the carrot into the mouth, limiting bite volume.




Regardless of the orientation or type of food on our fork, \REV{caregivers will} intuitively balance the bite volume \textit{efficiency} for a single bite with the \textit{comfort} of that bite. 
Motivated by this behavior, we present a bite transfer algorithm for selecting trajectories in a continuous space of mouth sizes, food geometries, and poses. Our approach (Section \ref{sec:approach}) takes as input a food mesh and an acquisition pose on the fork from the real world, and generates an analogous simulation environment. We learn a constraint model to sample goal food poses near the mouth, and perform motion planning based on a novel set of heuristics (Section~\ref{sec:heuristics}) to shape the perceived comfort and bite volume efficiency of each transfer.
To our knowledge, our approach is the first to formulate comfort and efficiency for bite transfer, to consider non-bite sized food items, and to work for a continuum of acquisition poses and food geometries. 
We demonstrate our algorithm in practice through a limited user study (Section \ref{sect:user_study}).
Our results show that while comfort alone and efficiency alone are able to outperform fixed trajectories on average, our approach of blending comfort and efficiency is the \textit{only} method to outperform a fixed pose baseline with statistical significance.
We run our method on various food items of differing geometries and scales in simulation (Appendix \ref{sec:experiments}). 





\section{Related Work}
\label{sect:related}
Our work draws inspiration not only from the state-of-the-art in the robot-assisted feeding literature but also from the shared autonomy and general robot-human handovers.

\smallskip
\noindent \textbf{Robot-assisted Feeding: Bite acquisition and transfer.}
Several specialized feeding devices for people with disabilities have come to market in the past decade. Although several automated feeding systems exist \cite{obi, myspoon, mealmate, mealbuddy}, they lack widespread acceptance as they use minimal autonomy, demanding a time-consuming food preparation process \cite{gemici2014learning}, or pre-cut packaged food and cannot adapt the bite transfer strategies to large variations due to pre-programmed movements. Existing autonomous robot-assisted feeding systems such as \cite{2019Feng}, \cite{2019Gallenberger}, \cite{2016Park}, and \cite{herlant_thesis} can acquire and feed a fixed set of food items, but it is not clear whether these systems can adapt to different food items that are either not bite-sized and require multiple bites or require other bite transfer strategies. Feng {\em et al.} \cite{2019Feng} and Gordon {\em et al.} \cite{gordoncontextual} developed an online learning framework using the SPANet network and showed \emph{acquisition} generalization to previously-unseen food items, but did not address the \emph{bite transfer} problem. Gallenberger {\em et al.} \cite{2019Gallenberger} showed a relationship exists between bite acquisition and transfer, but did not propose how to transfer bites for non bite-sized items in such a setting. Our paper aims to close this gap in bite transfer by developing a context-aware framework for robot-assistive feeding which generalizes to food items that are not bite-sized.

\smallskip
\noindent \textbf{Shared Autonomy for Robotic Assistance.}
Adding autonomy to provide robotic assistance to tasks by inferring human intent is a \REV{well-studied} field \cite{dragan2013policy, losey2020controlling, aronson2018eye, gopinath2016human, huang2016anticipatory, javdani2018shared, muelling2017autonomy,li2020multi}. This is especially relevant for precise manipulation tasks such as bite acquisition or bite transfer during robot-assisted feeding, while drawing parallels to other tasks such as peg-in-hole insertion \cite{lee2019making, unhelkarenabling}. For example, there has been work on using the concept of shared autonomy for bite acquisition tasks such as stabbing a bite, scooping in icing, or dipping in rice \cite{jeon2020shared}, where the researchers combined embeddings from a learned latent action space with robotic teleoperation to provide assistance. Unlike this body of work, this paper focuses on completely autonomous bite transfer of food items by keeping in mind our end user population, which may have severe mobility limitations.

\smallskip
\noindent \textbf{Robot-human Handovers.}
There are many works analyzing robot-human handovers, but most of the studies focus on objects that are handed over without using an intermediate tool \cite{agah1997humanInteractionRobot, edsinger2007cooperativeManipulation, huber2008handingOverTasks} in a single attempt. In this paper, we focus on tool-mediated handover of food-items that may not be bite-sized, and thus may require multiple handover attempts. The feeding handover situation poses an additional challenge of transferring to a constrained mouth, instead of a hand \cite{2019Gallenberger}. Gallenberger et al. \cite{2019Gallenberger} explore the problem of bite-transfer by providing the insight that bite transfer depends on bite acquisition and thus the transfer trajectories are not only food-item dependent but are also based on how a food was acquired. Cakmak et al. \cite{cakmak2011human} study the handover problem in an application-agnostic way, where they identify human preferences for object orientations and grasp types. Similarly, Aleotti et al. \cite{aleotti2012comfortableHandover} confirmed that orienting items in specific ways can make handover easier. Canal et al. \cite{gerard2016personalization} take a step further and explore how bite transfer can change with personal preferences. In our paper, we focus on tool-mediated bite-transfer of food items that may not be bite-sized and hence may require multiple transfer attempts.

\section{Context-Aware Multi-Bite Transfer}
\label{sec:approach}

\REV{A caregiver can guide a food item into their patient's mouth agnostic to the orientation of the food on the fork} --- they do not spend minutes optimizing how the food should be placed on the fork for the most optimal transfer. In this section, we first formalize the goal of \textit{food acquisition-agnostic} bite transfer, and then discuss our approach.

\subsection{Bite Transfer Problem Formulation}
\label{ssec:prob_formulation}

To begin bite transfer iteration, we are given a 3D mesh $\Mesh$ of the food item, the constant pose $\FoodPoseOnFork \in \mathbb{R}^6$ of the food item on the fork, a kinematics model for the robot and fork system with corresponding mesh $\RobotMesh$, and the pose estimate of the mouth $\MouthPose \in \mathbb{R}^6$. To capture the motion of the food item into the mouth, we want to find waypoints of the food item over time, represented by poses $\FoodPose \in \mathbb{R}^6$. Additionally, we assume the mouth can be represented by a simple elliptical tube $\MouthMesh$, where the ellipse axes are in the face plane, and open mouth dimensions $\MouthDims \in \mathbb{R}^2$ are specified per end user. These inputs are visualized in our PyBullet-based simulation environment in Fig.~\ref{fig:sim_overlap_and_sampling_timing}. We outline our method for acquiring these inputs in Appendix \ref{sec:rw_setup}.
Given these inputs, we formulate the goal of bite transfer as finding a sequence of food poses $\Traj = \{\FoodPose_0,\dots,\FoodPose_{L-1}\}$ of varying length $L$ respecting a set of physical constraints $\Constraints$ and cost function $\CT(\Traj)$, shown in Eq.~\eqref{eq:traj-objective}. 
For the task of bite transfer, $\Constraints$ consists of physical constraints. $\Constraints_0$ (Eq.~\eqref{eq:constraint0}) and \REV{$\Constraints_1$ (Eq.~\eqref{eq:constraint1})} ensure no collisions between the mouth mesh $\MouthMesh$ of dimensions $\MouthDims$ with pose $\MouthPose$ and the food mesh $\Mesh$ for each pose $\FoodPose_i \in \Traj$ as well as the robot-fork mesh $\RobotMesh$ respectively.
$\Constraints_2$ (Eq.~\eqref{eq:constraint2}) constrains the final food pose to be near the mouth opening, i.e., $\GoalFoodPose \doteq \FoodPose_{L-1}$ is in the support of goal pose distribution $\GoalDist$.
\vspace{2pt}
\begin{align*}
    \Traj^* &= \argmin_{\Traj} \CT(\Traj)
            \text{ s.t. } \Constraints_i(\Traj) = 1\ \forall \Constraints_i \in \Constraints \numberthis \label{eq:traj-objective} \\
    \Constraints_{0}(\Traj) &= \Mesh(\FoodPose_j) \cap \MouthMesh(\MouthPose) = \emptyset\ \forall \FoodPose_j \in \Traj  \numberthis \label{eq:constraint0} \\
    \Constraints_{1}(\Traj) &= \RobotMesh(\FoodPose_j) \cap \MouthMesh(\MouthPose) = \emptyset\ \forall \FoodPose_j \in \Traj  \numberthis \label{eq:constraint1} \\
    \Constraints_{2}(\Traj) &= \GoalFoodPose \in \GoalDist \numberthis \label{eq:constraint2} 
\end{align*}
\vspace{-15pt}

\subsection{Approach Overview}
\begin{figure}
    \centering
    \minipage{0.63\linewidth}
        \centering
        \includegraphics[width=.98\linewidth, trim={0 50pt 0 30pt}, clip]{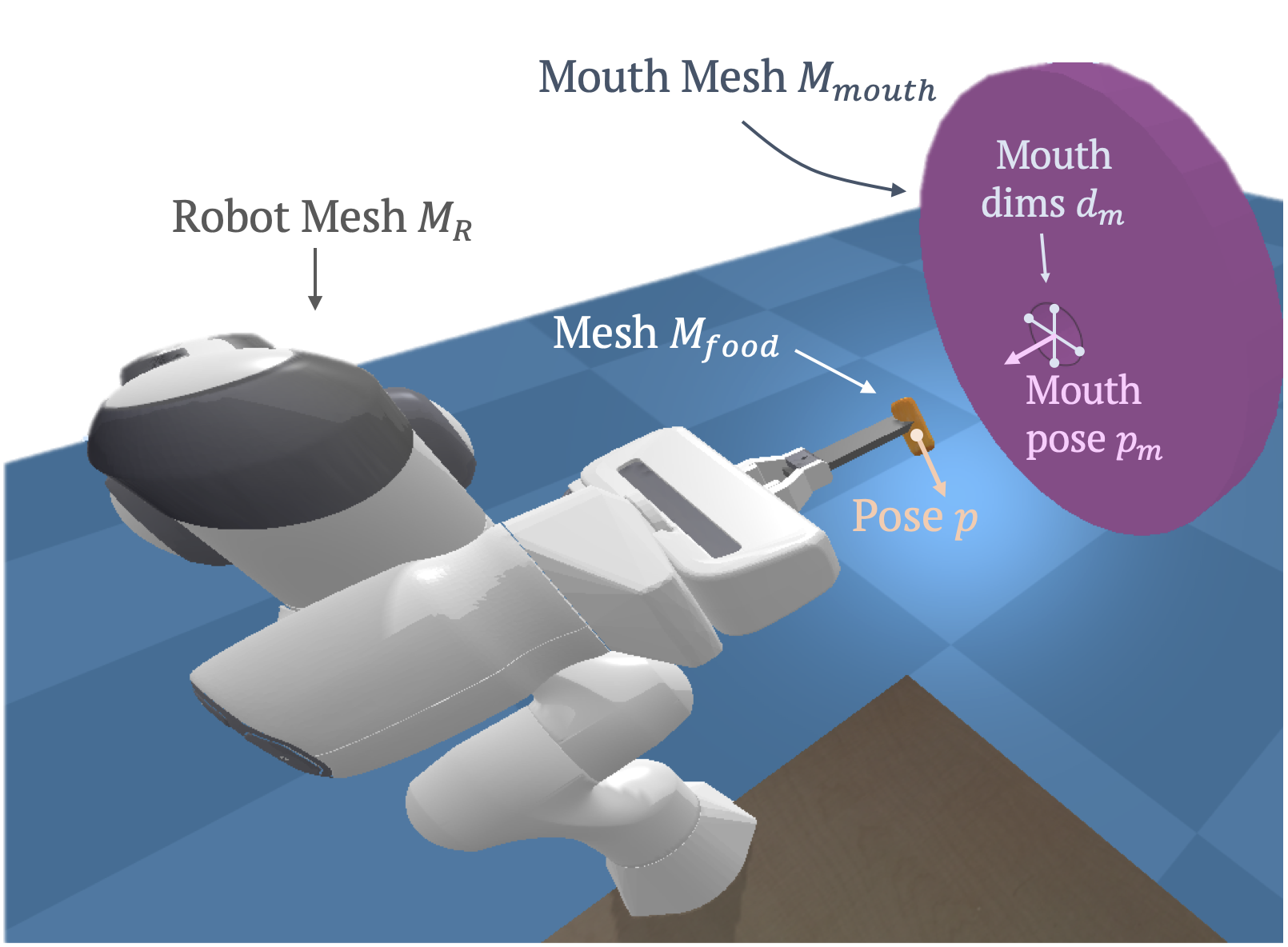}
    \endminipage%
    \hfill
    \minipage{0.35\linewidth}
        \centering
        \includegraphics[width=.99\linewidth, trim={410pt 5pt 0 5pt}, clip]{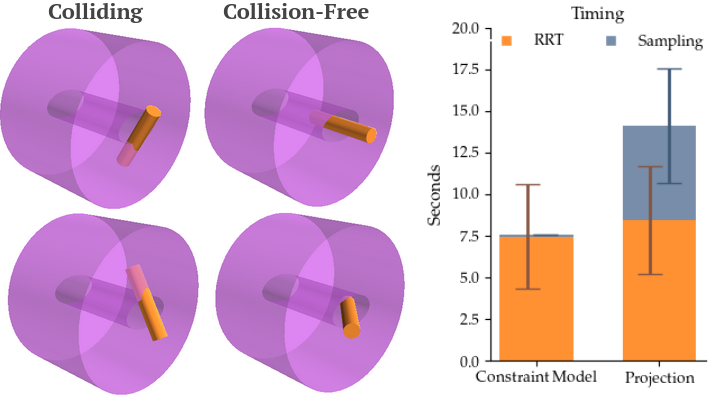}
        \captionsetup{width=.98\linewidth}
    \endminipage
    \vspace{-0.25cm}
    \caption{\textbf{Left:} PyBullet sim with robot mesh (Franka Emika Panda) $\RobotMesh$, food object mesh $\Mesh$ (e.g. carrot) at pose $\FoodPose$, and mouth mesh $\MouthMesh$ (cylindrical tube, radii from $\MouthDims$) at pose $\MouthPose$. \textbf{Right:} End-to-end algorithm timing for learned constraint model compared to projection-based sampling (100 trajectories each).}
    \label{fig:sim_overlap_and_sampling_timing}
    \vspace{-0.25cm}
\end{figure}

\REV{When taking a bite, a person intuitively \textit{simulates} the physics of their mouth's interaction with a carrot on our fork, regardless of the carrot's orientation, or where their arm starts. They might initially visualize where the carrot should be in the mouth and work backwards to find the most comfortable and efficient path.} Our approach captures this intuition. Our simulation environment (see Fig.~\ref{fig:sim_overlap_and_sampling_timing}) reflects the real world setup, where the mouth is replaced with a static elliptical mouth model, allowing us to simulate the interactions between the human mouth and the food item.

Our approach in Fig.~\ref{fig:front} consists of three phases: sampling, clustering, and planning. 
Since the space of feasible goal food poses in the mouth is continuous, we outline two efficient goal sampling methods (\emph{Projection} \& \emph{Learned Constraints}), which leverage simulation to batch sample from a set of ``feasible" orientations and offsets from the mouth, defined by the distribution $\GoalDist$, and then check these samples against the constraints $\Constraints$ to generate a varied set of feasible goal food poses $\GoalFoodPose$ near the mouth.
Next, we cluster the constraint satisfying poses into a set of $K$ goals with broad coverage over $\GoalDist$. We use heuristic guided bi-directional rapidly-exploring random trees to search for paths to goal food poses within the mouth that respect the physical constraints $\Constraints$. We guide the addition of new nodes to the h-BiRRT with a cost-to-come function $\CTC$ and a cost-so-far function $\CSF$, where the sum of $\CTC$ and $\CSF$ defines the overall predicted cost $\CT$ of a node in the h-BiRRT produced graph:
\begin{align}
    \CT(\Traj) \approx \CT(\{\FoodPose_0 \dots \FoodPose_i\}, \GoalFoodPose) = \CSF(\FoodPose_0 \dots \FoodPose_i) + \CTC(\FoodPose_i, \GoalFoodPose)
    \label{eq:sum_cost}
\end{align}

Section \ref{sec:heuristics} discusses how we incorporate comfort and efficiency into $\CTC$ and $\CSF$. Here, we first outline a method for generating goal poses $\GoalFoodPose \sim \GoalDist$ to satisfy the constraints $\Constraints$.


\smallskip
\noindent \textbf{Sampling Food Objects with Projection.}
When sampling goal food poses, there are certain fork orientations that are impossible or unsafe for the arm to reach (e.g. the fork pointing backwards relative to the face). We restrict the orientations of the robot end effector to be within a spherical cut centered on the into-mouth axis, and position offsets from the mouth center are bounded. These bounds form the uniform goal distribution $\GoalDist$. 
The full sampling algorithm is outlined in the appendix in Algorithm \ref{alg:sampling}. We first generate batches of food goal poses from $\GoalDist$ and check for collision, repeating this process until reaching $N$ collision-free samples or timing out.
Since a person's true mouth cavity fits within the tube-like elliptical mouth in simulation, which has a constant cross section in the mouth plane (see Fig.~\ref{fig:sim_overlap_and_sampling_timing}), we accelerate the 3D collision check by slicing the food by the mouth plane and then projecting the inner food mesh vertices for each goal pose onto the mouth plane (\emph{Projection}). The second image from the left in Fig.~\ref{fig:slice} (Appendix~\ref{sec:sampling_algs}) shows the slicing plane, with a sample carrot geometry. Finally we can check if the vertices are within the 2D mouth cross section to detect if the goal pose is collision-free. 

\smallskip
\noindent \textbf{Improved Sampling via Learned Constraints}. While \emph{Projection} checks samples for collision faster than a na\"ive 3D collision check, it still has significant and high variance lag (Fig.~\ref{fig:sim_overlap_and_sampling_timing}). We thus propose a sampling method, \emph{Learned Constraints}, that learns to predict constraint values (e.g., collision prediction) from 1M sim samples, with model inputs ($\MouthDims$, $\Mesh$, $\FoodPoseOnFork$, $\GoalFoodPose$). In Fig.~\ref{fig:sim_overlap_and_sampling_timing}, the right plot shows that a learned collision predictor significantly reduces sampling time. In Appendix \ref{sec:sampling_algs}, \REV{we provide further details and} show that \emph{Learned Constraints} maintain sample quality (e.g., predictive accuracy) and end-to-end trajectory performance (e.g., comfort \& efficiency costs).

\smallskip
\noindent \textbf{Clustering Goal Food Poses.}
Once we have timed out or reached $N$ collision free samples, we consolidate these goal poses into a representative set over $\GoalDist$ for the planning step.
We use a standard implementation of k-mediods, although any mediod clustering algorithm can be substituted.

\smallskip
\noindent \textbf{Motion Planning with Heuristic-Guided BiRRT.}
Once collision-free goal food poses have been generated and clustered, we must find trajectories to reach these goals. We adapt Rapidly-exploring Random Trees (RRT) for our motion planning. Inspired by Lavalle et al.~\cite{kuffner2000rrt}, who used bi-directional search ideas to grow two RRTs, we used one tree from the start state $\FoodPose_0$ and the other from the goal state $\GoalFoodPose^k$. To bias the two search trees towards each other, we take a heuristics-based approach \cite{urmson2003approaches}.
See Appendix \ref{sec:motion_planning} for details on our heuristic-guided implementation. Designing the heuristic cost functions will be discussed next.
\vspace{-1pt}

\section{Comfort \& Efficiency in Motion Planning}
\label{sec:heuristics}

The solution space of feasible transfer trajectories is often large: a small strawberry can be eaten in a wide variety of fork orientations due to its size and inherent symmetry.
Our approach narrows this solution space with \textit{comfort} and \textit{efficiency} heuristics during motion planning. One intuitive formulation is the path cost $\CSF$ being the distance between food poses (Eq.~\eqref{eq:dist_edge}), with the heuristic $\CTC$ being the distance to the goal pose (Eq.~\eqref{eq:dist_heuristics}).

\vspace{-15pt}
\begin{align}
    \CSF(\FoodPose_0...\FoodPose_i) &= \sum_{j=0}^{i-1}{||\FoodPose_{j+1} - \FoodPose_j||} \label{eq:dist_edge}\\
    \CTC(\FoodPose_i, \GoalFoodPose) &= ||\GoalFoodPose - \FoodPose_i|| \label{eq:dist_heuristics}
\end{align}

\begin{figure*}[t!]
    \centering
    
    \minipage{0.26\linewidth}
        \centering
        \includegraphics[width=.98\linewidth, trim={20pt 5pt 20pt 5pt}, clip]{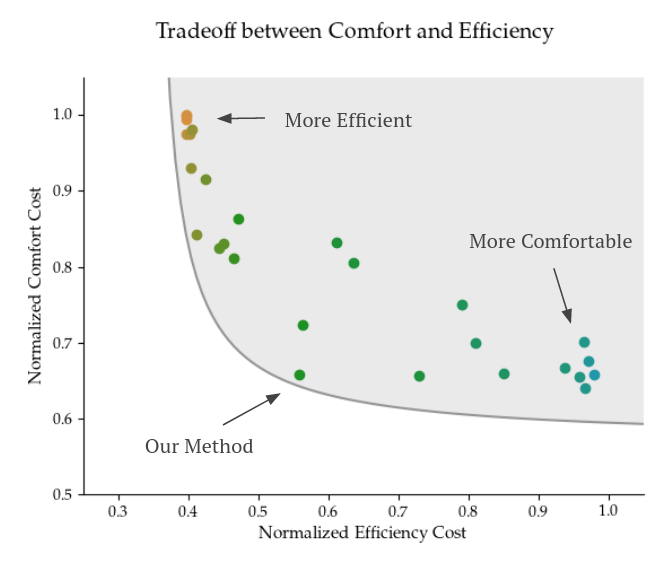}
    \endminipage%
    \hfill
    \minipage{0.73\linewidth}
        \centering
        \includegraphics[width=.94\textwidth, trim={10pt 5pt 30pt 5pt}, clip]{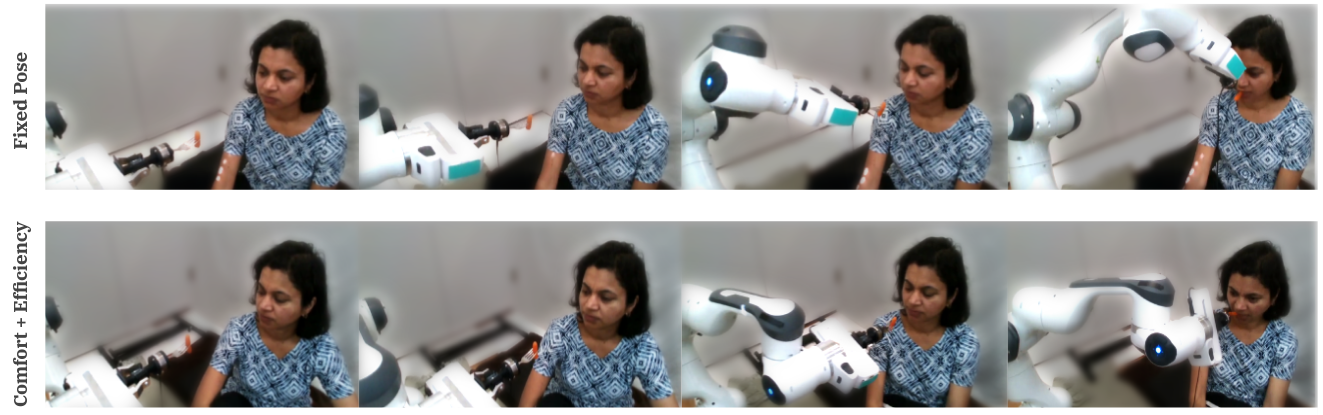}
    \endminipage%
    \vspace{-0.15cm}
    
    \caption{ \textbf{Left}: The fundamental trade off between average comfort and efficiency costs for a grid of chosen relative weightings of comfort and efficiency, with costs from our h-BiRRT method averaged over $500$+ initial food geometries and poses in simulation. Teal represents high ratios of comfort to efficiency, and orange the opposite. Our weights are the green dot at the elbow of this trade-off, achieving low efficiency cost and low comfort cost. \textbf{Right}: Sample trajectories for Fixed Pose (top) and Comfort+Efficiency (bottom) for the \Gvertical\ food geometry (see Section \ref{sect:user_study}). 
    While Fixed pose (baseline) incurs high comfort cost (close to user's face), our method finds a trajectory that is both comfortable and efficient for the user.
    }
    \label{fig:tradeoff_and_user_1_fixed_pose_vertical_example}
    \vspace{-0.35cm}
\end{figure*}

While this cost function guides hRRT to the goal pose, finding the shortest distance path in food pose space ignores both comfort and bite volume efficiency. Consider the vertically oriented carrot in the first row of Fig.~\ref{fig:tradeoff_and_user_1_fixed_pose_vertical_example}. If we sample a goal pose with the carrot oriented into the mouth and just past the teeth, a straight path to this goal is optimal in distance cost, but the person can barely take a bite; to add, the end effector would be close to the user's face, which could be considered uncomfortable. \REV{From our initial surveys with users with mobility limitations}, we indeed conclude that comfort and efficiency are essential during bite transfer. One participant comments, ``\emph{The orientation should be comfortable for the utensil and the food ... the arm should maintain a low profile to not obstruct sight ... it should be fast...}". 
To this end, we develop two competing cost functions to shape the trajectories produced by hRRT: (1) bite efficiency, capturing the percentage of the food inside the mouth at the end of a trajectory, and (2) trajectory comfort, capturing the perceived user comfort along a given trajectory.

\subsection{Modeling Efficiency}

The most \textit{efficient} goal pose brings the most food into a person's mouth, which can be measured in the real world by comparing the food mesh before and after each bite. In simulation, we approximate the new food mesh without knowing the biting physics for a user. Instead, we assume a bite slices the food mesh in the face plane (see Figure~\ref{fig:slice}). Let $V_i$ be the volume of the food geometry, and $V_f$ be the remaining volume after the bite.
We estimate the efficiency cost of goal poses with Eq.~\eqref{eq:efficiency}. The n-root ($n=3$ in practice) of the volume ratio amplifies the cost difference between goal poses of lower final volumes (high efficiency) to more noticeably bias RRT growth towards the most efficient goal poses. The resulting costs are in Eq.~\eqref{eq:efficiency_edge} \& \eqref{eq:efficiency_heuristics}.
\begin{align}
    \CT_E(\GoalFoodPose) &= (V_f / V_i)^{1/n}
    \label{eq:efficiency} \\
    \CSF(\FoodPose_0...\FoodPose_i) &= \sum_{j=0}^{i-1}{||\FoodPose_{j+1} - \FoodPose_j||} \label{eq:efficiency_edge} \\
    \CTC(\FoodPose_i, \GoalFoodPose) &= ||\GoalFoodPose - \FoodPose_i|| + \beta_E \CT_E(\GoalFoodPose) \label{eq:efficiency_heuristics}
\end{align}
Note that this cost only considers the goal pose, rather than the entire trajectory. We empirically found that other notions of efficiency applied to \textit{paths}, like trajectory execution time, or the distance of the path traveled, do not vary as much between outputs of h-BiRRT, and so yield less impact on the quality of trajectories produced.

\subsection{Modeling Comfort and Personal Space}

A trajectory that brings the arm too close within a user's \textit{personal space} could influence the user's perceived safety, even if the transfer efficiency is high. 
We develop a notion of comfort that draws from proxemics literature in human-robot interaction, a well-studied field \cite{takayama2009influences} for tasks such as in social robot navigation \cite{kirby2010social}. 
Building off the notion of ``personal space", we hypothesize that trajectories should stay within a conic region stemming from the mouth, with a wide cross-sectional area further from the face that narrows towards the mouth. Prior work in human factors for social navigation has shown that a person's comfortable personal space can be different for each cardinal direction, usually being larger within a person's visual field than outside \cite{kirby2010social}. Building on this intuition, we skew the cone down, away from the visual field.
\setlength{\columnsep}{5pt}
\begin{wrapfigure}{r}{0.55\linewidth}
    \vspace{-10pt}
    \centering
    \includegraphics[width=0.97\linewidth]{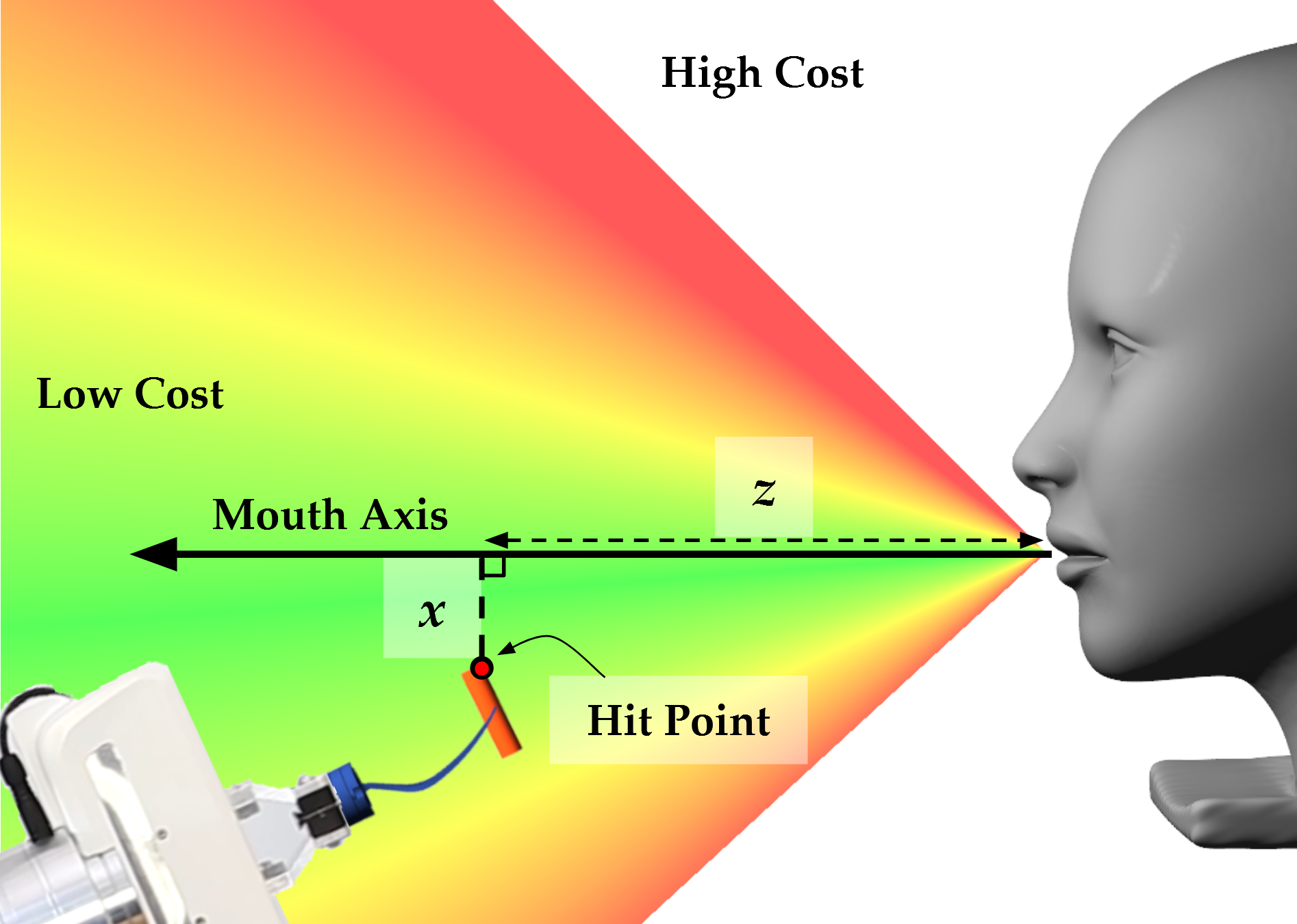}
    \caption{
    Spatial comfort cost (red higher, green lower). The steeper cost gradient in the upward direction than downward ensures trajectories near the face (e.g., Fig.~\ref{fig:tradeoff_and_user_1_fixed_pose_vertical_example}) have high ``comfort" cost. 
    }
    \label{fig:comfort}
    \vspace{-0.1cm}
\end{wrapfigure}
We define a spatial cost function resembling an elliptical Gaussian at each cross section centered along the mouth axis (Fig.~\ref{fig:comfort}). We posit that the upward direction relative to a person's face, which is closer to the visual field, should penalize deviation from the mouth axis more than in the downward direction. For a distance $z \in \mathbb{R}^+$ along the mouth axis and offset from the mouth axis $x \in \mathbb{R}^2$ in the cross section plane, we define the spatial comfort cost in Eq.~\eqref{eq:comfort}.
\begin{align}
    \CT_C^s(x, z) &= 1 - e^{-\alpha \frac{x^T\Sigma(x) x}{z^2}} \label{eq:comfort} \\
    \CT_C(\FoodPose) &= \frac{1}{NM}\sum_{h_j \in H}{\CT_C^{s}(h_j)} \label{eq:ray_reduce}
\end{align} 

Here, $\Sigma(x)$ is a piece-wise covariance matrix in the face plane. In our experiments, we used a diagonal covariance matrix, with smaller variances above the mouth horizontal plane than below, and equal variances left and right. This cost and the mouth axis can be visualized in Fig.~\ref{fig:comfort}.
Our comfort cost is applied on both the food item mesh and the entire simulated robot mesh and fork. In essence, we create a low resolution depth image from the perspective of the mouth and apply our cost function on the 3D location of each pixel. For a given food pose $\FoodPose$ and the corresponding simulated robot mesh, we cast rays in simulation along the mouth axis, starting at an $N \times M$ grid of points relative to the mouth center and on the face plane (e.g. $z = 0$), ending at a fixed maximum distance along the mouth axis $z_{\text{max}}$. The set of hit points from this ray cast, denoted $H = \{h_j \in \mathbb{R}^3\}$, are passed into the cost function in Eq.~\eqref{eq:comfort} and normalized by the total number of points (Eq.~\eqref{eq:ray_reduce}). This comfort cost is incorporated as a distance-weighted edge cost in hRRT with weight $\gamma_C$, and can be included in the heuristic as an additional goal cost using weighting $\beta_C$ (Eq.~\eqref{eq:comfort_edge} \& \eqref{eq:comfort_heuristics}).
\begin{align}
    \CSF(\FoodPose_0...\FoodPose_i) &= \sum_{j=0}^{i-1}{||p_{j+1} - \FoodPose_j|| \cdot (1 + \gamma_C \CT_C(\FoodPose_j, \FoodPose_{j+1}))} \label{eq:comfort_edge} \\
    \CTC(\FoodPose_i, \GoalFoodPose) &= ||\GoalFoodPose - \FoodPose_i|| + \beta_C \CT_C(\GoalFoodPose) \label{eq:comfort_heuristics}
\end{align} 
Here, $\CT_C(\FoodPose_j, \FoodPose_{j+1})$ is shorthand for the comfort cost at the midpoint of these two food poses. We denote this formulation as ``comfort only," since there is no consideration of efficiency here. Incorporating comfort alone can yield trajectories that keep the robot within the cone comfort region, but often this generates final goal poses that would not be easy to bite. Next, we will discuss incorporating both efficiency and comfort as costs for h-BiRRT.
\vspace{-2pt}

\subsection{Trading off Comfort and Efficiency}


Ideally, an assistive robot would be able to feed bites of food with both comfort \textit{and} efficiency in mind. In order to maximize both comfort and efficiency, we can put together the comfort costs (Eq.~\eqref{eq:comfort_edge} \& \eqref{eq:comfort_heuristics}) and efficiency costs (Eq.~\eqref{eq:efficiency_edge} \& \eqref{eq:efficiency_heuristics}), yielding the cost functions for h-BiRRT in Eq.~\eqref{eq:comf_and_eff_edge} \& \eqref{eq:comf_and_eff_heuristics}, where the weightings $\beta_E$, $\beta_C$, and $\gamma_C$ emphasize the efficiency at the goal, comfort at the goal, and comfort along the trajectory, respectively.
\begin{align}
    \CSF(\FoodPose_0...\FoodPose_i) &= \sum_{j=0}^{i-1}{||\FoodPose_{j+1} - \FoodPose_j|| \cdot (1 + \gamma_C \CT_C(\FoodPose_j, \FoodPose_{j+1}))} \label{eq:comf_and_eff_edge} \\
    \CTC(\FoodPose_i, \GoalFoodPose) &= ||\GoalFoodPose - \FoodPose_i|| + \beta_C \CT_C(\GoalFoodPose) + \beta_E \CT_E(\GoalFoodPose) \label{eq:comf_and_eff_heuristics}
\end{align} 
In Fig.~\ref{fig:tradeoff_and_user_1_fixed_pose_vertical_example}, we plot the average comfort and efficiency scores for our h-BiRRT pipeline over a grid of weight values for $\beta_E$, $\beta_C$, and $\gamma_C$ and over a large number of initial food poses and geometries (e.g., carrots, strawberries, celery, cantaloupes) in simulation. Refer to Appendix~\ref{sec:experiments} for quantitative results and example trajectories for each food type. Optimizing for efficiency only finds trajectories with the highest comfort costs but lowest efficiency costs, and vice versa for comfort only. This demonstrates that there is in fact a trade-off in comfort and efficiency costs when running h-BiRRT with our heuristic functions. Our approach will choose the ``elbow" of this trade-off, balancing both efficiency and comfort. 

\section{User Study}
\label{sect:user_study}


\begin{figure*}
    \centering
    \includegraphics[width=0.96\textwidth, height=0.15\textwidth]{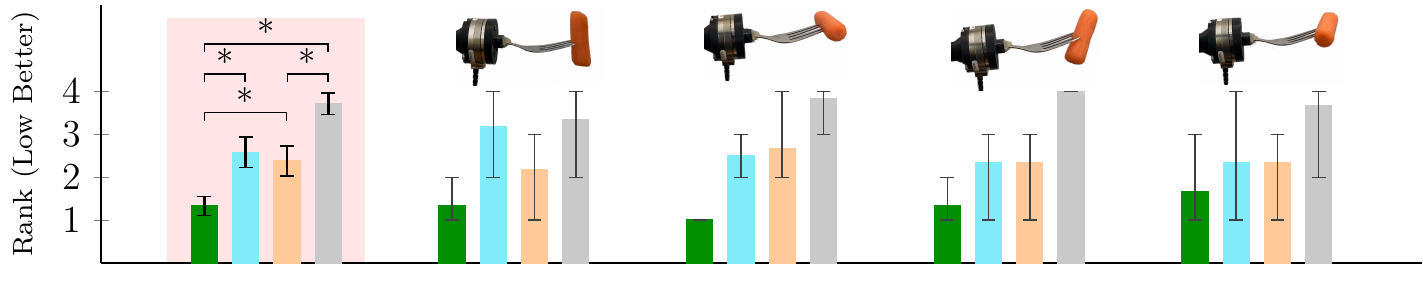}\vspace{-0.25cm}
    \vspace{-0.25cm}
    \includegraphics[width=0.957\textwidth, height=0.14\textwidth]{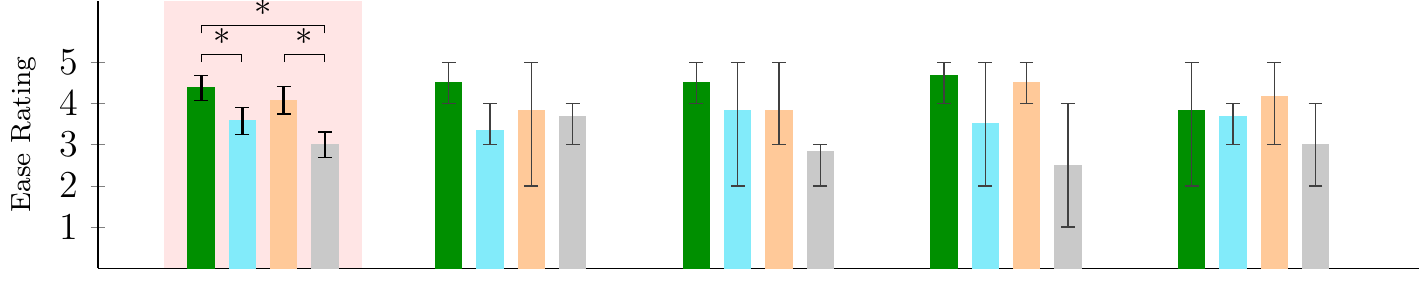}
    \includegraphics[width=0.945\textwidth, height=0.185\textwidth]{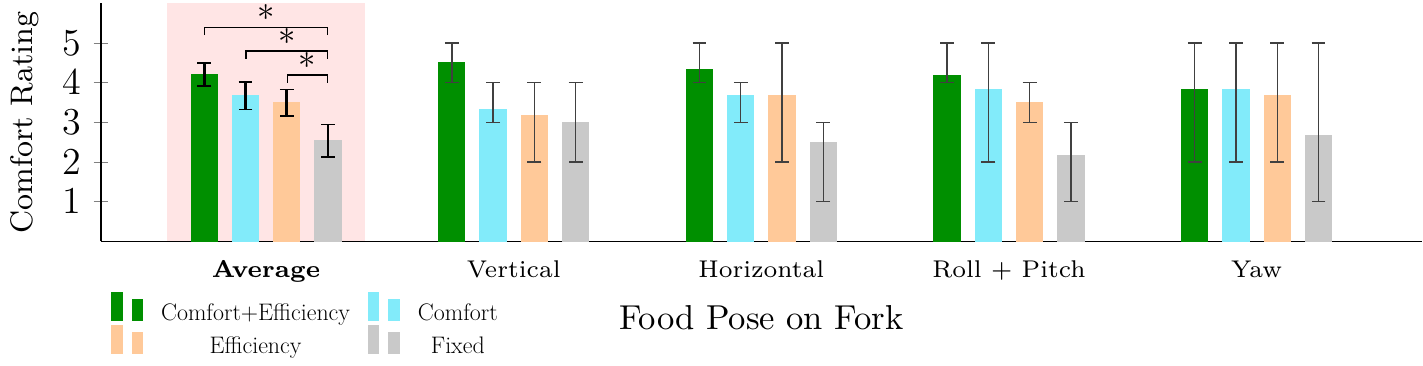}
    \vspace{-0.1cm}
    \caption{User study quantitative results. Each non-highlighted plot shows the average comfort rating, ease rating, and rank between trajectory types across 4 different food poses, with the range across all 6 users plotted as error bars. The highlighted plots on the left show the average across all food poses, with error bars representing 95\% confidence. \REV{Significant results, as determined by two-way ANOVA with repeated measures, Tukey HSD test, and Bonferroni correction ($P < 0.01$), are marked with an asterisk.} We do not treat multiple ratings as independent. Despite the limited sample size ($N=6$), trajectories from the combined comfort and efficiency method perform significantly better than the baseline fixed pose approach across all three metrics. Notably, the efficiency-only method often performs worse than comfort-only in comfort ratings. \REV{See Appendix \ref{sec:user_study_appendix} for significance testing details and more analysis.}
    }
    \label{fig:user_results}
    \vspace{-0.4cm}
\end{figure*}

\subsection{Experimental Setup}

\begin{figure*}[b]
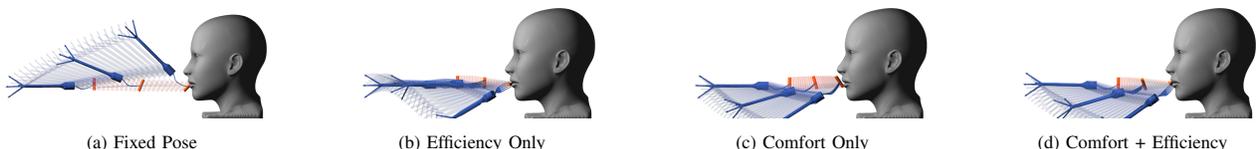

    \vspace{-15pt}
    \centering
    \begin{subfigure}[b]{0.24\textwidth}
         \centering
         \includegraphics[width=0.95\textwidth, trim={5pt 30pt 5pt 20pt}, clip]{figures/fixed_pose.png}
         \caption{Fixed Pose}
     \end{subfigure}
     \begin{subfigure}[b]{0.24\textwidth}
         \centering
         \includegraphics[width=\textwidth, trim={5pt 30pt 5pt 20pt}, clip]{figures/efficiency.png}
         \caption{Efficiency Only}
     \end{subfigure}
     \begin{subfigure}[b]{0.24\textwidth}
         \centering
         \includegraphics[width=\textwidth, trim={5pt 30pt 5pt 20pt}, clip]{figures/comfort.png}
         \caption{Comfort Only}
     \end{subfigure}
     \begin{subfigure}[b]{0.24\textwidth}
         \centering
         \includegraphics[width=\textwidth, trim={5pt 30pt 5pt 20pt}, clip]{figures/ce.png}
         \caption{Comfort + Efficiency}
     \end{subfigure}
    \vspace{0.1cm}
    \caption{Example trajectories optimized for each metric. The efficiency metric (b) rotates the carrot sideways so as much can be consumed in one bite as possible. The comfort metric (c) penalizes more complicated trajectories where the robot body is likely to encroach on the face. The combined metric (d) results in a fairly straight trajectory that still ends with a sideways carrot.}
    \label{fig:trajectory_visualizations}
\end{figure*}

We conducted a user study with six \REV{non-disabled} participants to evaluate the perceived comfort and efficiency with our real world setup \footnote{\REV{We decided to recruit non-disabled participants due to Covid-19 \& safety concerns.}
Please see Appendix \ref{sec:user_study_appendix} for further discussion.}
In Appendix \ref{sec:rw_setup}, we discuss our real world system design, and how we ensure user safety during our user studies. 
Key parameter choices are shown in Table \ref{tab:exp_params} in the Appendix. 
We consider carrots of varying sizes and fixed acquisition poses, visualized in the first row of Fig.~\ref{fig:user_results}: 
\Gvertical, 
\Ghorizontal, 
\Grollleftpitchup, and
\Gyawleftforward.
Users were instructed to sit still facing the robot, and to take a bite of each food item after each trajectory if they felt comfortable to do so. In addition, an emergency stop button was placed next to them for added assurance. See Appendix~\ref{sec:user_study_appendix} for more user study details. We evaluated the following methods:

\setenumerate{noitemsep}
\begin{enumerate} [leftmargin=1.5em]
    \item \Bfixed\ (F): We fix the final orientation of the food item independent of the pose of the food item on the fork \REV{and} the food size. This final orientation is hard-coded for a specific type of food and is inspired by the taxonomy of food manipulation strategies developed in \cite{bhattacharjee2018food}.
    \item \Beffonly\ (E): Our approach with h-BiRRT and only efficiency costs, Eq.~\eqref{eq:efficiency_edge} and \eqref{eq:efficiency_heuristics}.
    \item \Bcomfonly\ (C): Our approach with h-BiRRT and only comfort costs, Eq.~\eqref{eq:comfort_edge} and \eqref{eq:comfort_heuristics}.
    \item \Bours\ (CE): Our approach with both efficiency and comfort in mind: the h-BiRRT cost functions use both efficiency and reward, Eq.~\eqref{eq:comf_and_eff_edge} and \eqref{eq:comf_and_eff_heuristics}.
\end{enumerate}

For each food pose and method, we evaluate two trajectories end-to-end with each user.
After two trajectories for a given method, we ask a series of questions to gauge the user's perceived comfort of each trajectory, and the ease with which they were able to take a bite. We compare responses to these questions, in terms of
\Mratings\ (the average user rating of \textit{comfort} for each evaluated trajectory, from 1 to 5, with 5 being the most comfortable), 
\Mease\ (average rating of their \textit{ease} of taking a bite for each evaluated trajectory, normalized from 1 to 5, with 5 being the best), 
\Mrank\ (Average relative rank of each method, from 1 to 4), and 
\Msafety\ (Average user rating of safety, from 1 to 5).

\subsection{Results}

The ease rating, comfort rating, and approach rank are summarized in Fig.~\ref{fig:user_results}. Importantly, our real world evaluation pipeline (Appendix \ref{sec:rw_setup}) was perceived as safe to the user regardless of food geometry or method, achieving an average \Msafety \ rating of 4/5.
Despite the limited sample size, our method (CE) significantly outperforms the fixed baseline (F) for all three metrics (example in Fig.~\ref{fig:tradeoff_and_user_1_fixed_pose_vertical_example}), and consistently outperforms comfort-only (C) and efficiency-only (E), significantly so in Rank ratings.
Additionally, efficiency-only (E) did not perform as well as comfort-only in Comfort ratings.
This supports our hypothesized connection between user comfort perception and our comfort model (Eq.~\eqref{eq:comfort}).

The data is consistent with our hypothesis that, while optimizing over individual metrics (C and E) provides some improvement over the baseline, joint optimization performs even better in creating trajectories robust to real-world variation. We suspect that this is due to the large space of possible trajectories, where maximizing for only comfort puts no guarantee on efficiency, and vice versa. 


Qualitatively, our comfort model's sensitivity to objects above mouth level fits with user expectations. When asked about low-ranked trajectories, users stated that they believed ``\emph{the robot should have approached from underneath,}" or that they ``\emph{didn't like when [the robot] came up close to [their] face.}"
Users were more likely to instinctively move backwards when approached from above, near the face, and lean in when approached from below. In Fig.~\ref{fig:trajectory_visualizations}, we visualize a sample real world trajectory produced by each method for a \Gvertical \ pose. \Bfixed \ is neither maximally efficient (carrot only partially fits in mouth) nor comfortable (robot is too close to face). Common quantitative metrics like time and path length are not as informative in gauging comfort, so we limit our evaluation to these qualitative metrics. Appendix~\ref{sec:user_study_appendix} elaborates on quantitative and qualitative metrics and outlines how our approach naturally extends to the multi-bite setting with sample real world evaluations.

\section{Discussion} 
\label{sect:discussion}

\noindent \textbf{Summary.}
We present \REV{an approach based on motion planning} for bite transfer under a continuous space of possible acquisition angles. During planning, we narrow down the solution space of possible trajectories into the mouth with an awareness of both bite \textit{efficiency} and user \textit{comfort}. 
Our user study demonstrates that considering comfort and efficiency jointly provides \textit{significantly} more preferable trajectories compared to a fixed pose baseline. Furthermore, our method with comfort and efficiency consistently outperforms considering only comfort or only efficiency.

\smallskip
\noindent \textbf{Limitations.}
One limitation of our method is the assumption that the mouth can be represented by a rigid elliptical tube, and that the food item is also rigid. In reality, the human mouth and the food item can both be deformable, which expands the set of ``collision-free" paths into the mouth.


Furthermore, our user study only involved six non-disabled users due to Covid-19 related policies. In future work we plan to evaluate with more users, including users with mobility-impairment disabilities. However, we are excited that even with the given sample size, our method improves on the state-of-the-art with statistical significance.

\section*{Acknowledgements}
This work is funded by NSF Award Numbers 2132847 and 2006388, and by the Office of Naval Research.








\newpage
\bibliographystyle{IEEEtran}
\bibliography{IEEEabrv,references}

\clearpage
\newpage

\appendices

Additional details for our approach are described in the following sections. Appendix~\ref{sec:sampling_algs} covers the \emph{Projection} and \emph{Learned Constraints} sampling approaches in greater detail, including quantitative comparisons. Appendix~\ref{sec:motion_planning} provides details for the h-BiRRT algorithm used for motion planning, for example how the cost heuristics in Section~\ref{sec:heuristics} are incorporated into RRT growth with key parameter choices. Appendix~\ref{sec:rw_setup} explains our real world robotic pipeline for autonomous feeding, involving gravity compensation, food geometry perception, trajectory following, force-reactive control, and facial keypoint tracking. Appendix~\ref{sec:experiments} outlines simulation experiments, with quantitative and qualitative information to supplement our real world analysis. Finally, user study details including demographics and survey questions are covered in Appendix~\ref{sec:user_study_appendix} along with a discussion of how our algorithm naturally extends to the multi-bite setting using examples from our user study.

\section{Sampling Algorithms}
\label{sec:sampling_algs}

\begin{figure}[h!]
    \minipage{0.48\textwidth}
    \renewcommand\algorithmiccomment[1]{\hfill $\triangleright$ #1}
    \begin{algorithm}[H]
        \small
        \begin{algorithmic}[1]
        \floatname{algorithm}{Procedure}
        \STATE{Given mesh $\Mesh$, Constraints $\Constraints$, Goal sampling distribution $\GoalDist$, Mouth $\MouthDims$ and $\MouthPose$, Clustering method \ClusterFn}
        \STATE{$P = \{\}$}
        \WHILE{$P$ has less than $N$ poses}
            \STATE{Sample $B$ goal poses $\{\GoalFoodPose^{(1)} \dots \GoalFoodPose^{(B)}\} \sim \GoalDist$} \label{alg:sampling:line:sample}
            \STATE{$V \leftarrow$ vertices for $\Mesh$ at each pose in sample batch}
            \STATE{$V_{\text{sl}} \leftarrow$ SliceIfInside($V$, $\MouthPose$, $\MouthDims$)} \label{alg:sampling:line:slice}
            \STATE{$V_{\text{pr}} \leftarrow$ Projection($V_{\text{sl}}$, $\MouthPose$)} \label{alg:sampling:line:proj}
            \STATE{Add $\GoalFoodPose^{(i)}$ to $P$ if vertices $V_{\text{pr}}^{(i)}$ are inside mouth ($\MouthDims$)} \label{alg:sampling:line:coll}
            
        \ENDWHILE
        \RETURN{cluster centers $\leftarrow$ \ClusterFn$(P)$}
        \end{algorithmic}
        \caption{Projection Goal Pose Sampling}
        \label{alg:sampling}
    \end{algorithm}
    \endminipage
    \hfill
    \minipage{0.48\textwidth}
    \centering
    \includegraphics[width=\textwidth]{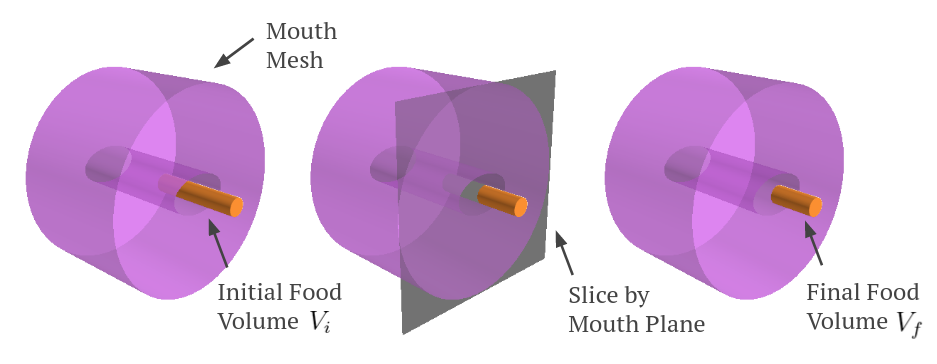}
    \caption{An example of what ``slicing" a food geometry, in this case a carrot, with the mouth plane, the gray cross sectional plane, looks like in simulation when calculating the efficiency score with simplified mouth and carrot geometries. The efficiency cost $\CT_E(\GoalFoodPose)$, defined in Eq.~\eqref{eq:efficiency}, uses the ratio of final volume $V_f$ and initial volume $V_i$, shown here on the right and left sides, respectively.}
    \label{fig:slice}
    \endminipage
\end{figure}

Our projection sampling algorithm (Algorithm \ref{alg:sampling}) begins by generating batches of food goal pose samples from $\GoalDist$ (Line~\ref{alg:sampling:line:sample}), and performing collision checks on each food pose with the mouth (Lines~\ref{alg:sampling:line:slice},~\ref{alg:sampling:line:proj},~\ref{alg:sampling:line:coll}), and repeating this process until reaching $N$ valid samples or timing out. We observed that performing 3D collision checks was the time bottleneck for each sampling iteration. After reaching $N$ samples, finding the representative goal set can be accomplished by any clustering algorithm that returns cluster centers from within the original set of food poses, since it may be the case that cluster centers outside of this set do not satisfy the same collision checks. The clustering method is referenced in the final line of Algorithm \ref{alg:sampling}. $\GoalDist$ is defined by fixed positional offsets from the mouth center and a spherical cut of feasible orientations for the fork. As an aside, while fixed positional offsets were sufficient for most food geometries, we noticed that using these fixed 3D offsets with very large geometries or small mouths occasionally leads to a sparse set collision-free poses, since these geometries may need more ``wiggle" room near the mouth to generate even a few collision free samples. One could compensate for this by adjusting this 3D offset as a function of the food geometry bounding box.

\begin{figure*}[h]
    \centering
    \minipage{0.33\linewidth}
        \centering
        \includegraphics[width=\linewidth]{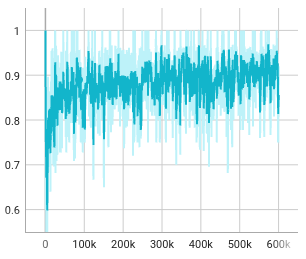}
    \endminipage
    \hfill
    \minipage{0.33\linewidth}
        \centering
        \includegraphics[width=\linewidth]{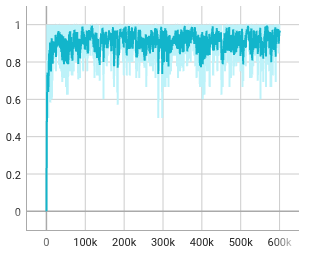}
    \endminipage
    \hfill
    \minipage{0.33\linewidth}
        \centering
        \includegraphics[width=\linewidth]{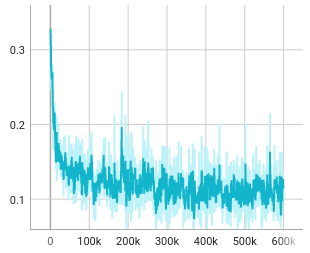}
    \endminipage
    \caption{Holdout training plots for the learned collision prediction model. From left to right: Holdout collision prediction accuracy during training for colliding samples (0 to 1); Holdout collision prediction accuracy during training for non-colliding samples (0 to 1); BCE holdout loss.}
    \label{fig:training_curves}
\end{figure*}

We further improve upon the efficiency of sampling with a learned constraint model, replacing lines \ref{alg:sampling:line:slice} \& \ref{alg:sampling:line:proj} with a forward pass of the constraint model and filtering by the outputs. As outlined in Fig.~\ref{fig:sim_overlap_and_sampling_timing} in the text, sampling with a learned model greatly improves the speed of the full algorithm. We use a 3-layer NN to predict collisions from food pose on fork $\FoodPoseOnFork$, food mesh $\Mesh$, mouth dimensions $\MouthDims$, and a desired goal food pose $\GoalFoodPose$. \REV{We trained on 1 million samples of just the final pose of the robot and the corresponding simulated constraint label. Our data collection in simulation involved variations in not just food type and initial pose on the fork, but also in food scaling in all 3 axes and in mouth elliptical cross section dimensions. We chose the bounds for each of these factors to be a super set of those likely to be encountered in the real world. Note that each of these variables is passed in as input to the constraint model and are accessible in the real world. Thus, this model can be readily used in real world settings that allow access to these state variables. The main sim-to-real gap here involves the elliptical mouth assumption, which we hope to tackle in future work. }

\smallskip
\REV{During training, we accounted for label shift by weighting constraint satisfying and non-constraint satisfying samples proportionally to their occurrence in the simulation dataset. Figure~\ref{fig:training_curves} shows the holdout prediction accuracies and cross entropy loss during the training process.} Table~\ref{tab:coll_acc} shows that the constraint predictor is accurate at a slightly lower rate than the projection sampling method (i.e. high \textit{sample quality}); however, the faster sampling time for the constraint model allows for more iterations of sampling in order to offset the lower prediction accuracy (although this fact is not leveraged for Table~\ref{tab:coll_acc}). \REV{Due to the imperfect predictions, the \emph{Learned Constraints} model on its own may result in collision generating goal poses after clustering. To prevent these poses from being evaluated, we run an additional constraint check on the cluster centers to filter out any prediction errors with no loss in final goal pose quality.}
\begin{table}
    \begin{center}
        \begin{tabular}{c|cc}
            & \emph{Learned Constraints} & \emph{Projection} \\
            \hline
            \rule{0pt}{2ex}    
            Sampling (s) & $0.015 \pm 0.032$ & $5.6 \pm 3.5$ \\
            Constraint Acc. (\%) & $74 \pm 13$ & $82 \pm 11$ \\
            RRT Comfort Cost & $0.63 \pm 0.26$ & $0.61 \pm 0.26$ \\
            RRT Efficiency Cost & $0.56 \pm 0.21$ & $0.54 \pm 0.24$ \\
        \end{tabular}
        \caption{\emph{Learned Constraints} vs. \emph{Projection} sampling over 100 evaluated trajectories. $(\mu \pm \sigma)$ for sampling time, constraint prediction accuracy, RRT comfort cost (normalized), RRT efficiency cost (normalized).}
        \label{tab:coll_acc}
    \end{center}
\end{table}

Additionally, we show that the final h-BiRRT generated trajectories achieve similar comfort and efficiency costs in simulation compared to the projection-based sampling method (i.e. high \textit{trajectory quality}). In summary, the learned constraint sampling achieves similar quality trajectories while greatly reducing the algorithm execution time.

\section{Motion Planning Algorithms}
\label{sec:motion_planning}

\begin{algorithm}[H]
    \small
    \begin{algorithmic}[1]
    \floatname{algorithm}{Procedure}
    \STATE{Given $K$ goal poses $\{\GoalFoodPose^k\}$, cost-to-come $\CTC$, cost-so-far $\CSF$, initial food pose $\FoodPose_0$, sampling distribution $\TrajDist$, hRRT \textit{connect}}
    \STATE{Start tree: $S = \text{Tree}(p_0)$}
    \STATE{Goal trees: $G = \{G_k = \text{Tree}(\GoalFoodPose^k)\}$}
    \WHILE{$S$ not connected to each $G_k$}
        \STATE{Sample unconnected goal tree $G_i \sim G$ by goal cost} \label{alg:hrrt:line:goal_tree_sampling}
        \STATE{$T_A, T_B \leftarrow $ sortByLength($S$, $G_i$)}
        \STATE{\textit{connect}$(T_A, T_B, \CSF, \CTC)$ \label{alg:hrrt:line:rrt_connect}} 
    \ENDWHILE
    \RETURN{Smoothed paths from $\FoodPose_0$ to each goal $\GoalFoodPose^k$}
    \end{algorithmic}
    \caption{Heuristic-Guided Bi-directional RRT (h-BiRRT)}
    \label{alg:hrrt}
\end{algorithm}

For our h-BiRRT implementation, outlined in Algorithm~\ref{alg:hrrt}, we sample goal poses as described previously, and produce a goal tree associated with each of the sampled goal poses (Line~\ref{alg:hrrt:line:goal_tree_sampling}). To bias the start to each goal tree, we leverage a heuristics-based approach \cite{urmson2003approaches} with a parameter $m_{q}$ that is used to weight the Voronoi region associated with each node, such that
\begin{align}
m_{q} = 1 - \frac{c_{i} - c^*}{c_{\text{max}} - c^*},
\label{eq:quality}
\end{align}
where $c_{i}$ is the total cost associated with node $i$, $c^*$ is the total cost of the optimal path from start to goal, and $c_{\text{max}}$ is the highest total cost of any node. We calculate total cost $c$ as the sum of cost-to-come $g$ and the heuristic cost-to-go $h$. Specifically for the goal tree, we use a simple distance heuristic from goal to start, and for the start tree, we use the novel comfort and efficiency heuristics.  For each tree, we use k-Nearest Neighbors to select nearest nodes in the tree to expand. At every iteration, we run the bi-directional RRT planner for the same start tree, but different sampled goal trees, where the probability of sampling a goal tree at any iteration is proportional to its goal cost (Lines~\ref{alg:hrrt:line:goal_tree_sampling},~\ref{alg:hrrt:line:rrt_connect}). This ensures that we eventually reach every goal tree from the start while prioritizing reaching the least costly goal pose first. 

\smallskip
\noindent \textbf{Picking a trajectory.}
Once hRRT has completed, the resulting tree has a smoothed and low cost path to each of the $K$ goal poses that were generated from sampling. Each of these trajectories has an associated cost defined by evaluating $\CSF$ on the entire trajectory, and adding any costs associated with that trajectory's goal pose. The final trajectory that will be evaluated on the real robot arm is the trajectory with the lowest overall cost.

\smallskip
\noindent \textbf{Key Parameter Choices.}
Our key cost parameters are shown below. $N$ is the number of collision free samples to generate before clustering; $K$ is the number of clusters generated after sampling; $n$ is power to which the bite efficiency ratio (final over initial) is raised in the efficiency cost; $\alpha$, $r_{\text{up}}$, $r_{\text{down}}$, $r_{\text{side}}$ parameterize the comfort cost; $\beta_E$ is the goal efficiency contribution; $\beta_C$ is the goal comfort contribution; and $\gamma_C$ is the edge comfort contribution.

\begin{table}[H]
    \centering
    \begin{tabular}{ccccccccc}
        $N$ & $K$ & $n$ & $\alpha$ & $r_{\text{up}}$ & $r_{\text{down}}$, $r_{\text{side}}$ & $\beta_E$ & $\beta_C$ & $\gamma_C$ \\
        \hline
        \rule{0pt}{2ex}    
        150 & 15 & 3 & 1.0 & 1.5 & 1. & 1. & 10. & 10. \\
    \end{tabular}
    \caption{Key parameters for our method.}
    \label{tab:exp_params}
\end{table}

\section{Robotic System for Autonomous Feeding}
\label{sec:rw_setup}

In order for an assistive feeding platform to be adopted, users need to feel safe while our algorithm is operating in the real world. In the context of assistive feeding, the autonomous robot must execute a complex end-effector trajectory produced by our method and simultaneously react to the highly sensitive human mouth without causing discomfort. We designed a robotic system to evaluate expressive trajectories with an awareness of users and the forces they exert on the utensil.

\begin{figure}
    \centering
    \includegraphics[width=0.48\textwidth]{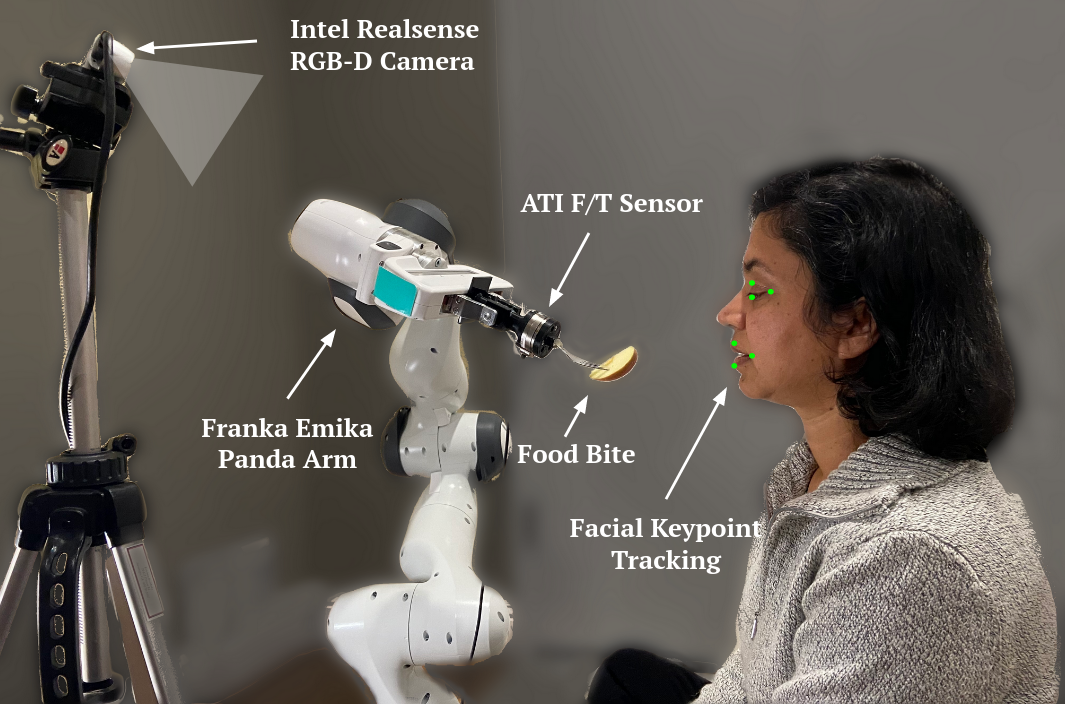}
    \caption{Our real world evaluation setup consists of a Franka Emika Panda robotic arm, a fork attached to an ATI Mini45 6-axis F/T Sensor via a custom 3D printed mount, and an external Intel Realsense RGB-D camera. The F/T sensor enables the robot arm to interact with the user's mouth through a force-reactive controller. The external camera enables food geometry perception at the start of each bite transfer and visual servoing based on facial keypoints during trajectory execution.}
    \label{fig:rw_setup_annotated}
\end{figure}

The trajectory execution operates as follows. At initialization, the force/torque readings from the end-effector are decoupled from gravity forces/torques of the unknown food object by estimating the food mass from a sequence of end effector poses. Using an RGB-D camera, we estimate the 3D mesh of the food item on the fork. Once our algorithm produces a sequence of waypoints for the food item, we smooth the trajectory and interpolate through the waypoints using bounded end effector velocities. Along the trajectory, force and torque readings along with robot state are utilized by a force-reactive admittance controller to safely navigate the forces applied by users' mouths in end-effector space. We delve into each of these components in the following sections, and our real world setup is shown in Fig.~\ref{fig:rw_setup_annotated}.


\smallskip
\noindent \textbf{Food Mass Gravity Compensation.}
We use an ATI F/T 6-axis Mini45 sensor (see Fig.~\ref{fig:rw_setup_annotated}) to receive force and torque readings from a custom fork mounted on the robot's end-effector. Before compensation, these readings have an unknown bias and do not take gravity into account. At the beginning of every episode, we cycle through end-effector poses to estimate the inherent bias of the force torque sensor readings along with the food mass currently on the fork. Note that this procedure is not specific to feeding and can be extended to robotic systems with unknown sensor biases and initial gravity forces.

Upon acquiring a new bite of food, the end-effector cycles through several pose quaternions $\poi \in \mathbb{R}^4$ for $i \in 1 \dots N$ and records the raw force and torque readings $\fri \in \mathbb{R}^3$ and $\tri \in \mathbb{R}^3$ at each pose. Let $\Tes \in \mathbb{R}^{3x3}$ be the fixed known rotational transform from the robot end-effector coordinate frame to the sensor coordinate frame, and let $\Twe \in \mathbb{R}^{3x3}$ be the known rotational transform from the world frame to the robot's current end-effector orientation $\poi$. Additionally, let the force sensor bias be $\fbi$ and the torque bias be $\tbi$. Assuming an unknown mass $m$ and a known torque radius $r$ (the distance to the tip of the fork) in the end-effector frame +y direction, and a fixed \& known gravity in the world frame -z direction $g$, we know that, 
\begin{align}
    \fri + \fbi &= \Tes \Twe m \begin{bmatrix} 0 \\ 0 \\ -g \end{bmatrix} \label{eq:force_comp}\\
    \tri + \tbi &= \Tes \left(\begin{bmatrix} 0 \\ r \\ 0 \end{bmatrix} \times  \Twe m \begin{bmatrix} 0 \\ 0 \\ -g \end{bmatrix}\right) \label{eq:torque_comp}
\end{align}

Here, $m$, $\fbi$, and $\tbi$ yield 7 unknown scalars. Note that Eq.~\eqref{eq:force_comp} and \eqref{eq:torque_comp} are linear in these unknowns, and can thus be rearranged into the form $\mathbf{Ax} = \mathbf{b}$ with $\mathbf{x} \in \mathbb{R}^7$. We then set up a system of linear equations for each robot pose $\poi$, and the corresponding force measurements. With multiple poses, we can solve the least squares approximation to estimate the unknowns in the presence of sensor noise. With the F/T readings properly adjusted, we can proceed to estimating the 3D geometry of the food item.

\begin{figure}
    \centering
    \minipage{0.5\linewidth}
    \includegraphics[width=\linewidth]{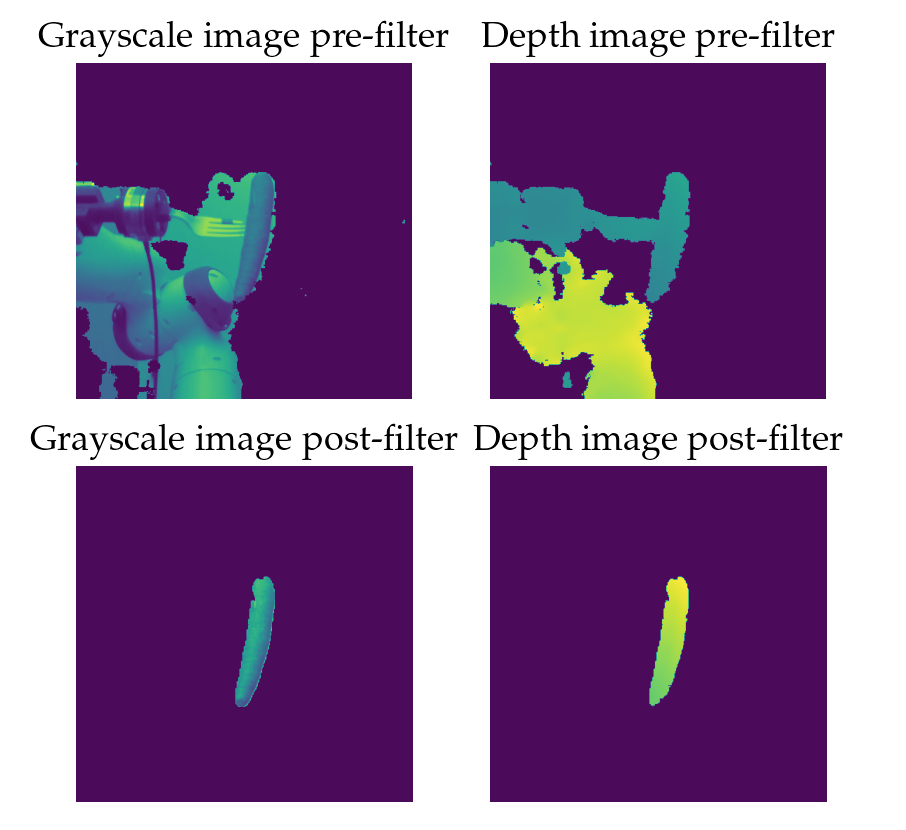}
    \endminipage
    \minipage{0.5\linewidth}
    \includegraphics[width=\linewidth]{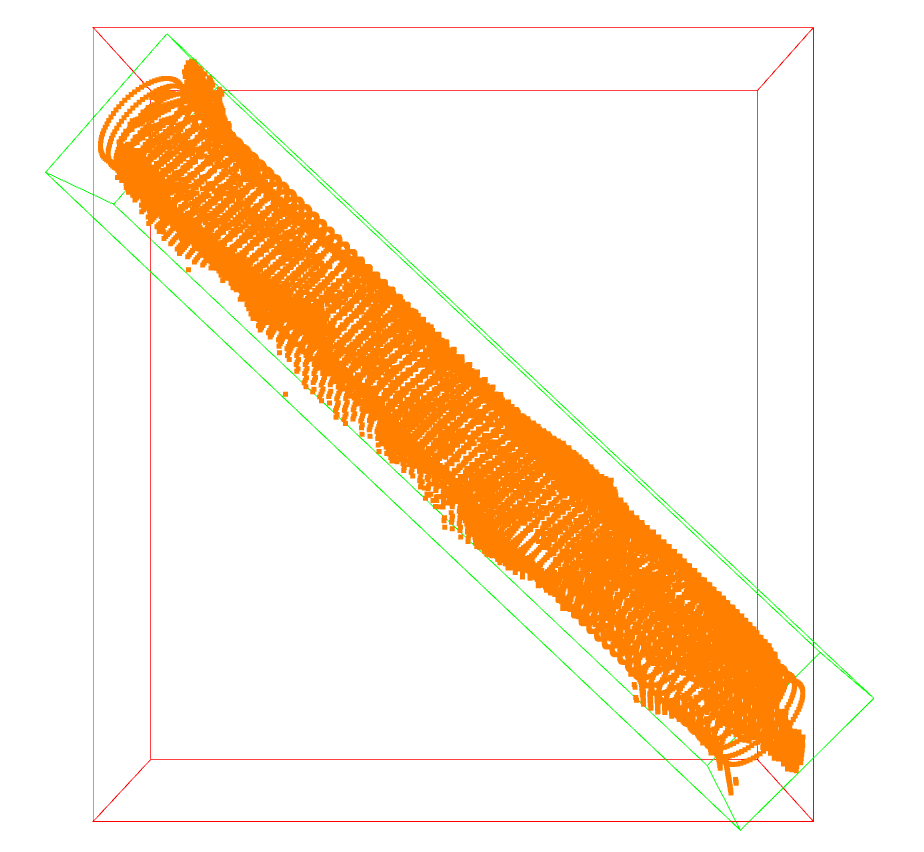}
    \endminipage
    \caption{\REV{Left: HSV + Depth filtering example for a carrot. Right: Point cloud reconstruction example, with estimated oriented bounding box in green.}}
    \label{fig:pcd}
\end{figure}

\noindent \textbf{Food Perception.}
Food shape can vary greatly between different food items, even of the same food type. In the multiple bite formulation, this food geometry can change over time, and so 3D geometry must be re-evaluated after each bite in order to obtain the most accurate simulation environment for searching for trajectories into the mouth. 
Our food perception pipeline is as follows: we use a fixed externally mounted RGB-D camera to take several images of the food item for multiple fork orientations. \REV{We assume we can extract the pixel region of the food item. For the purposes of our study, where food items have a single color, we extract the food region using HSV and depth filtering, but any segmentation model can be used to extract the food item from each image.} Then we use the RGB-D image to reconstruct a point cloud for each image, and we stitch these pointclouds together with knowledge of the camera frame pose relative to the robot frame of reference (see Fig.~\ref{fig:pcd}).

\begin{table*}[h!]
    \centering
    \begin{tabular}{c|cccc}
        & $\beta_C\text{-},\ \gamma_C\text{-},\ \beta_E\text{-}$ & $\beta_C\uparrow,\ \gamma_C\uparrow,\ \beta_E\downarrow$ & $\beta_C\downarrow,\ \gamma_C\downarrow,\ \beta_E\uparrow$ & \REV{Dist-Only}\\
        \hline
        \rule{0pt}{2ex}
        RRT Comfort Cost & $0.65 \pm 0.25$ & $0.62 \pm 0.15$ & $0.97 \pm 0.22$ & $0.79 \pm 0.21$ \\
        RRT Efficiency Cost & $0.56 \pm 0.20$ & $0.98 \pm 0.13$ & $0.43 \pm 0.18$ & $0.91 \pm 0.15$\\
    \end{tabular}
    \vspace{0.3cm}
    \caption{Final performance (lower is better) over 500+ evaluated sim trajectories for three cost regimes over cost weights $\beta_C$, $\beta_E$, and $\gamma_C$, relative to our chosen values. $(\mu \pm \sigma)$ for h-BiRRT produced comfort cost (normalized) and efficiency cost (normalized). From left to right: our algorithm, at the elbow of the trade-off of comfort and efficiency; high comfort weighting regime, with low comfort cost but high efficiency cost; high efficiency weighting regime, with low efficiency cost but high comfort cost; \REV{distance only regime, with no comfort or efficiency heuristics.}}
    \label{tab:sim_quant}
\end{table*}

\setlength{\columnsep}{10pt}
\smallskip
For the purposes of demonstrating our approach, our food geometry estimation procedure assumes that the food item can be sufficiently represented by a bounding box or cylinder, seen in Fig.~\ref{fig:pcd}, although in the future we hope to make this pipeline more robust as well as capable of representing more complex 3D mesh geometries. Importantly, overestimating the bounding region of a food item, e.g. with a convex hull, does not hinder our ability to produce valid collision free trajectories. The expressiveness of trajectories produced by our method scales with the accuracy of the mesh geometries we get from the real world.

\smallskip
\noindent \textbf{Trajectory Following.}
Once we have recovered the food geometry and pose on the fork, our algorithm samples a set of valid goal poses of the food item, and then finds trajectories to each goal pose as discussed in Sections \ref{sec:approach} and \ref{sec:heuristics}, in the form of discrete waypoints to the goal pose. In order to ensure user comfort and consistency, we fix the velocity norm in end effector space along the trajectory. We smooth the output of the h-BiRRT, and generate an interpolated waypoint for each time step of trajectory evaluation bounded by the max velocity norm. We proceed to the next waypoint when the end effector comes within a ``follow" radius. We found that using a proportional (P) controller to generate velocities along the interpolated trajectory was sufficient to stay within the follow radius.

Despite moving at relatively slow speeds, the human mouth is very sensitive to any forces being applied on it, especially once the fork has entered the mouth. We implemented a simple force-reactive controller that filters velocities from the trajectory following controller, discussed in the next section.

\smallskip
\noindent \textbf{Force-Reactive Controller.}
The biased and gravity compensated F/T readings are brought into the robot's frame of reference. We use a force admittance controller to convert forces and torques experienced by the fork to linear and angular velocity commands for the robot. In practice, we used only linear velocity commands from the force readings, yielding the following control equations for the end effector velocities:
\begin{align}
    v(t) = K_p e(t) + K_i\int_0^t{e(\tau)d\tau} + K_d \frac{de(t)}{dt}  \label{eq:pid_ft} \\
    e(t) = \min(F(t) + F_{th}\ ,\ \max(F(t) - F_{th}\ ,\ 0))
\end{align}

This PID controller aims to keep the forces applied by the fork within a band $\{-F_{\text{th}}, F_{\text{th}}\}$ to prevent causing discomfort to the user. This threshold is determined by the sensitivity of the human mouth. In practice we set this to be around $0.25$ N. If the forces are outside of this band, the algorithm ignores the velocity command from the trajectory following algorithm, and instead returns the force-reactive controller command.

\smallskip
\noindent \textbf{Facial Keypoint Visual Servoing.}
Another important aspect of ensuring user comfort is knowing the position of human mouth in 3D space relative to the robot arm. One common approach is to mount a camera on the end effector of the robot, and track keypoints on a person's face using facial detection algorithms. However, our method leverages a high degree of mobility at the end effector, meaning the human mouth or face may not always be fully in view or at an upright orientation. To ensure consistent facial tracking, we mount an RGB-D camera at an external position, which measures the 3D position of the person's head and mouth. Our trajectory waypoints are adjusted relative to this known 3D mouth position, but velocities are computed in the global frame to ensure the motion is smooth and of bounded velocity.

\section{Simulation Experiments}
\label{sec:experiments}

\begin{figure*}
    \centering
    \includegraphics[width=\textwidth]{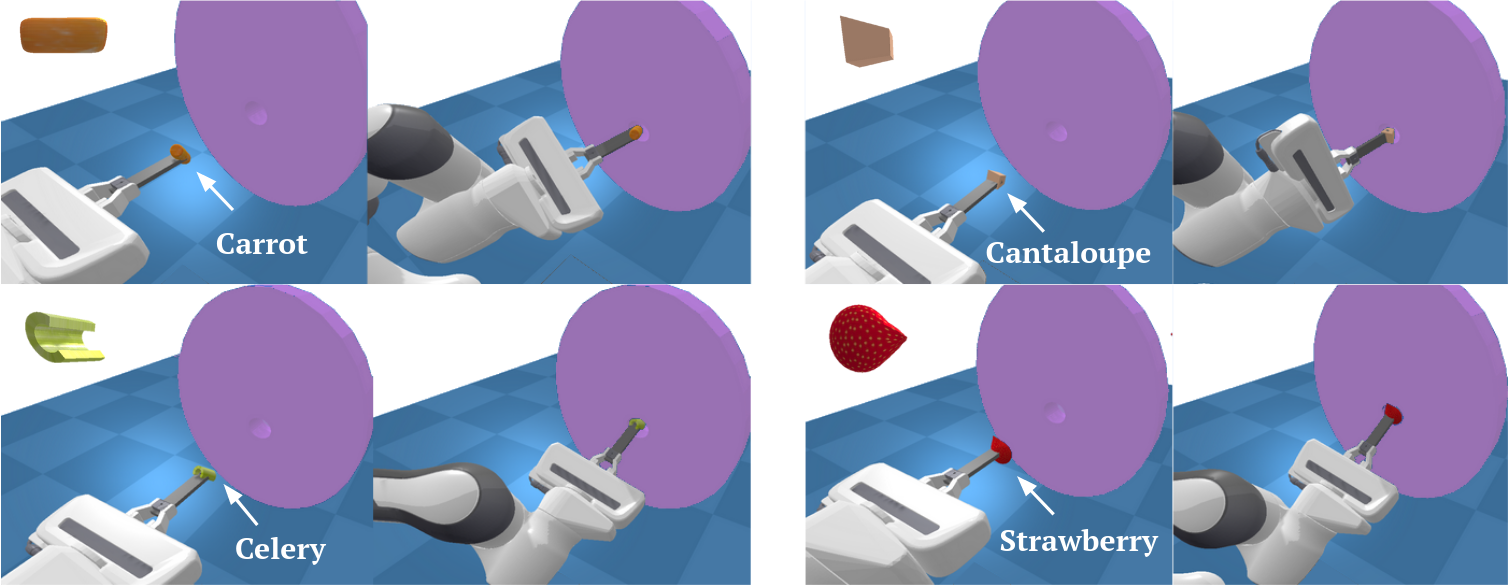}
    \caption{Starting and ending poses for our method (CE) evaluated in simulation with models for many different basic geometries: a carrot (top left), slice of cantaloupe (top right), celery (bottom left), and strawberry (bottom right). Given the specific food item's mesh $\Mesh$, our method is able to find paths that bring a large portion of each food item into the mouth while simultaneously respecting the user's personal space.}
    \label{fig:different_food_items_sim}
\end{figure*}


\smallskip 
\noindent \textbf{Different Food Geometries.}
To illustrate our flexibility in regards to food geometries, we model four geometry types of random size in simulation that each have a unique 3D shape: a carrot (cylinder); a piece of cantaloupe (trapezoidal prism); a stalk of celery (half of a cylindrical tube); and a strawberry (``teardrop" shape). These geometries are visualized in Fig.~\ref{fig:different_food_items_sim}. The initial and final poses for a sample trajectory under our method for each food type are also shown in this figure. Notably, our algorithm is able to bring each food item at least partially into the mouth, picking a final goal pose and planning a trajectory that is both comfortable and efficient. The trade-off plots from Fig.~\ref{fig:tradeoff_and_user_1_fixed_pose_vertical_example} in the text were generated using these food geometries of various scales, and thus represent a super-set of the generated trajectories in the user study experiments in Section~\ref{sect:user_study}, which considers carrots alone.

\smallskip
\noindent \textbf{Comfort and Efficiency.} 
For a wide range of values of $\beta_E$, $\beta_C$, and $\gamma_C$, we ran $500$+ trajectories in simulation for a range of initial food poses, types (see above), and geometry sizes, recording the average comfort cost, efficiency cost, and distance costs. Fig.~\ref{fig:tradeoff_and_user_1_fixed_pose_vertical_example}, discussed previously, illustrates the results of this parameter sweep, and Table~\ref{tab:sim_quant} provides the cost values for this sweep under our weights (elbow of trade-off), high relative comfort weighting, and high relative efficiency weighting. As our intuition suggests, we find that there exists a trade-off between comfort and efficiency under our formulation. Since respecting a user's personal space can yield end effector orientations that bring the food into the mouth with low efficiency, we note that trajectories that are more \textit{comfortable} on average have lower \textit{efficiency}, and vice versa. \REV{Additionally, distance-only heuristics produce notably higher comfort and efficiency costs, suggesting that only using the distance heuristic produces trajectories far from the Pareto front of comfort and efficiency}.

\section{User Study}
\label{sec:user_study_appendix}

\smallskip
\noindent \textbf{Design.}
Our user study involved six non-disabled participants, (ages 19, 22, 23, 23, 53, \& 55, three women and three men). Three had experience interacting with robots, and three did not. All had experience feeding other people. \REV{Due to COVID-19 \& safety concerns, we were unable to recruit people with motor impairment into our user study. We have been limited even in the number and population of non-disabled users over the past year. Additionally, we believe the problem of assistive feeding is important beyond its impact on users with disabilities. Specifically, real-world evaluation with non-disabled users still captures the complex interactions between the mouth and the robot, and provides insights about formalizing comfort for the general public. Similar comfort formalisms might even extend to other real-world human-robot applications, as well as to assistive feeding beyond the disabled population.} 

Per user, we ran two iterations for each method (\Bfixed, \Bcomfonly, \Beffonly, \Bours), all for each of the carrot food poses:
\begin{enumerate}[leftmargin=1.5em]
    \item \Gvertical: 
    the carrot begins along the up-down axis, perpendicular to the fork.
    \item \Ghorizontal: 
    the carrot begins along the left-right axis, perpendicular to the fork.
    \item \Grollleftpitchup: and
    the orientation includes a roll angle left and pitch angle up.
    \item \Gyawleftforward:
    the orientation is rotated about the vertical axis with respect to \Ghorizontal.
\end{enumerate}

The size of each carrot we used varied between each experiment, and we biased the \Gyawleftforward\ category towards bigger carrots for added diversity. For sanitation, the forks were cleaned after every two iterations, and food items were replaced if not eaten. The following scale questions were used to evaluate the user's perception of comfort after every two iterations (i.e. for one method and food geometry combination):
\begin{enumerate}[leftmargin=1.5em]
    \item \Mease: I was able to easily take a bite (1 to 5).
    \item \Mratings: I was able to comfortably take a bite (1 to 5).
    \item \Msafety: I felt safe while being fed (1 to 5).
\end{enumerate}

After each method for a given food geometry (eight total iterations), the user was asked to rank the methods in order of preference. Additionally, they were asked to comment on how they would change the robot trajectories if they were feeding someone who cannot feed themselves. Finally, they were given the opportunity to provide additional comments about the methods \& robot system.

\begin{figure*}
    \centering
    \includegraphics[width=0.98\textwidth]{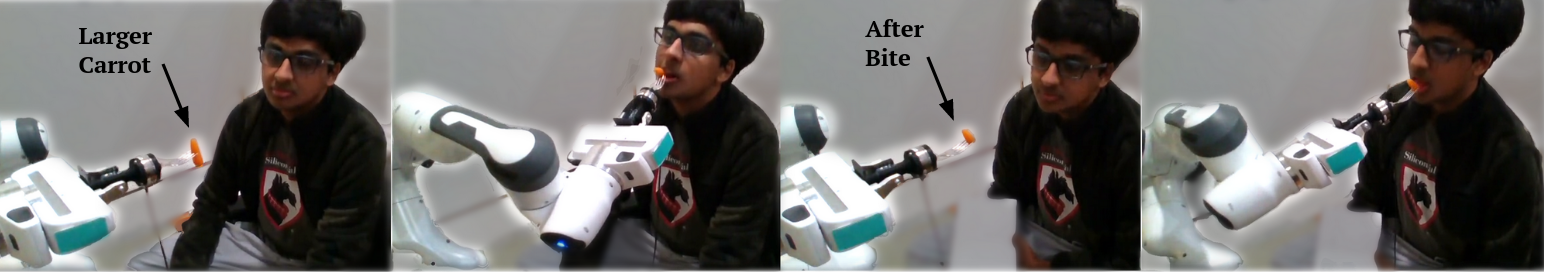}
    \caption{An example of a food geometry that is too big to comfortably fit in the mouth in a single iteration. Our method (CE) was evaluated on this larger carrot over multiple iterations. Our food perception re-evaluates the 3D geometry at each feeding iteration (the first and third images from the left), allowing our algorithm to find new trajectories at each feeding iteration that balance making progress at each bite (efficiency) and respecting user personal space (comfort). Two iterations were required using our method, and the final pose of each trajectory is shown in the second and fourth images from the left.}
    \label{fig:multi-bite}
\end{figure*}

\smallskip
\noindent \REV{\textbf{Objective Metrics}.
We considered several objective metrics before choosing to use subjective metrics for our evaluation. We will now discuss the shortcomings of various objective metric options: elapsed time, path length, trajectory jerk, applied force, \& vital data. We find that elapsed time depends more on the distance of travel for the robot than the difficulty of taking a bite. The path length, in turn, depends only on the change in joint state from the initial to goal robot configurations, and therefore both elapsed time and path length vary in predictable and uninformative ways. Specifically, path traversal time varied from 9 seconds to 17 seconds for each transfer, while bite time remained roughly constant at 1-1.5 seconds. For trajectory jerk, our planned trajectories are all quite smooth by design, so this metric is equally uninformative. Applied force is also not informative since each bite taken induces a significant impulse on the fork, often noticeably larger than the minor interaction forces that might make transfer uncomfortable. Vital data would be quite interesting to analyze, but we believe that subjective questionnaires are the most informative avenue for gauging perceived comfort, as prior work has also noted. Further complicating many of these metrics, users might decide to not take a bite of the food during the study, for example if the presented orientation is too difficult. Thus, these objective metrics would fail to capture the user’s real opinions about these trajectories, whereas subjective metrics can be informative in all cases.}

\smallskip
\noindent \REV{\textbf{Safety Considerations}.
In order to ensure the user's safety, we took several precautions during the experiment, in addition to those described in Appendix \ref{sec:rw_setup}. The user sits facing the robot. The proctor sits on the left side of the table from the user's perspective. The emergency-stop button is placed either in the hands of the user, or next to them on a side table, depending on the user's preference. A fork (a cut-out of a standard metal table fork) is mounted via a custom 3D mount on the force/torque sensor, which is in turn attached to a shaft that the Franka Panda gripper can hold. Importantly, the fork mechanism is not rigidly attached to the Panda arm, and thus allows the fork to interact with the mouth much more forgivingly. In addition to the force reactive control which runs throughout the experiment, this mechanism is designed to break from the gripper when too much force is applied (although this scenario was never encountered). This experiment was conducted under IRB-52441.}

\smallskip
\noindent \REV{\textbf{Results}.
Due to the measurement of 3 different variables (Rank, Ease, and Comfort), we apply a Bonferroni correction to our significance threshold for all hypotheses ($P < 0.01$). First, given two independent variables (Food Pose and Method), we analyzed our data with a Two-Way ANOVA with repeated measures. This determined that Method alone had a significant effect on Rank, Comfort, and Ease ($P < 0.001$). We therefore averaged out the Food Pose and analyzed Method vs (Rank, Comfort, Ease) with a One-Way ANOVA with correlated samples. Significant results (as determined by a Tukey HSD Test, $P < 0.01$) are marked in Figure \ref{fig:user_results} with an asterisk. For the reader's convenience, the significant pairings are listed in Table \ref{tab:listed_sig}.
}

\begin{table}[h!]
    \centering
    \begin{tabular}{c|ccccc}
        \hline
        \rule{0pt}{2ex}
        Rank & CE vs C & CE vs E & CE vs F & C vs F & E vs F \\
        Ease & CE vs C & CE vs F & E vs F & & \\
        Comfort & CE vs F & C vs F & & & \\
        \hline
    \end{tabular}
    \vspace{0.3cm}
    \caption{\REV{Pairwise significance test results for each method, under each user rating metric. Methods are \Bours\ (CE), \Beffonly\ (E), \Bcomfonly\ (C), \& \Bfixed\ (F). A vs B indicates method A $>$ method B.}}
    \label{tab:listed_sig}
\end{table}

\REV{We did not treat multiple user ratings as independent data. We note that, while these results are significant despite our limited sample size ($N=6$), this does not necessarily mean that they generalize to the target population of individuals with limited upper-body mobility.}

\REV{Our results support the hypothesis that combining comfort and efficiency heuristics facilitated more comfortable bite transfer for non-disabled participants than either heuristic on their own. One interesting finding is that the Comfort-only model does not perform significantly better than the Efficiency-only model in Comfort ratings. We note that a user’s \textit{perceived} comfort is multi-faceted, and includes factors beyond just the trajectory taken for bite transfer. For example, forces applied on the mouth due to inefficient food orientations, jitter of the robot arm due to the reactive controller, and minor positioning errors in the real world can often play a role in user comfort. In this work, we attempt to formalize the trajectory-based component of comfort; we demonstrate that when combined with Efficiency heuristics, these Comfort heuristics perform better than either heuristic does on its own, often with statistical significance}

\smallskip
\noindent \textbf{Multi-Bite.} 
Often, a single bite of food might be too large or not oriented on the fork in the most optimal way. In these cases, it is necessary for intelligent bite transfer algorithms to \textit{recognize} that the food item cannot be consumed in a single bite and then \textit{plan} in a closed loop over each bite transfer iteration, continually re-planning based on how the food geometry changed after the last bite. Since our algorithm takes as input the full 3D mesh, our method is capable of dealing with the large changes in food geometry that can occur after each bite. A sample multi-bite transfer in the real world that uses our algorithm with both comfort and efficiency is shown in Fig.~\ref{fig:multi-bite}. When feeding themselves a carrot of this size, a person might take a \textit{greedy} approach to taking multiple bites, making as much progress as one can at every iteration without ever sacrificing one's comfort until the food item has been fully consumed --- while it might be possible to bring this entire carrot into the mouth in one bite (i.e. high efficiency), one can imagine this might not be comfortable for the user. Our algorithm plans over multiple bites in a similarly greedy fashion: we consider only the current food geometry, and then make a decision over what goal and subsequent trajectory to take to minimize our overall cost for the bite iteration. After the user takes a bite, we repeat this process. By incorporating both comfort and efficiency into our cost function, our approach naturally extends to this multi-bite planning process as shown with the sample trajectory in Fig.~\ref{fig:multi-bite}. As a result, our method required only two bites to feed the entire carrot on the fork and was able to keep the user comfortable throughout.

\end{document}

%% file: preamble.tex
\usepackage{algorithm}
\usepackage[noend]{algorithmic}
\usepackage{amsmath}
\usepackage{amssymb}
\usepackage{color}
\usepackage{xcolor}
\usepackage{graphicx}
\usepackage{dblfloatfix}
\usepackage{multicol}
\usepackage{multirow}
\usepackage{subcaption}
\usepackage{tabularx}
\usepackage{times}
\usepackage{siunitx}
\let\labelindent\relax
\usepackage{enumitem}
\usepackage{xurl}

\makeatletter
\let\NAT@parse\undefined
\makeatother
\usepackage{hyperref}
\hypersetup{
    colorlinks=true,
    linkcolor=orange,
    filecolor=magenta,
    urlcolor=orange,
    citecolor=orange,
}
\usepackage{pdfx}

\newcommand\numberthis{\addtocounter{equation}{1}\tag{\theequation}}

\DeclareMathOperator*{\argmin}{argmin}

\usepackage{tikz}
\usetikzlibrary{arrows,shapes,backgrounds,patterns,fadings,decorations.pathreplacing,decorations.pathmorphing}
\tikzset{>=stealth'}
\usetikzlibrary{arrows}
\usetikzlibrary{bayesnet}

\usepackage[skins]{tcolorbox} 

\newcommand{\Mesh}{\mathcal{M}_\text{food}}
\newcommand{\MouthMesh}{\mathcal{M}_\text{mouth}}
\newcommand{\RobotMesh}{\mathcal{M}_R}
\newcommand{\Traj}{\mathcal{T}}
\newcommand{\Constraints}{\mathcal{C}}
\newcommand{\GoalDist}{\mathcal{D}_g}
\newcommand{\TrajDist}{\mathcal{D}_{\Traj}}
\newcommand{\ClusterFn}{k-mediods}
\newcommand{\FoodPose}{p}
\newcommand{\FoodPoseOnFork}{p_f}
\newcommand{\GoalFoodPose}{p_G}
\newcommand{\MouthPose}{p_m}
\newcommand{\MouthDims}{d_m}

\newcommand{\CTC}{h}  
\newcommand{\CSF}{g}  
\newcommand{\CT}{\mathcal{J}}  

\newcommand{\fri}{\mathbf{f_i}}
\newcommand{\fbi}{\mathbf{f_b}}
\newcommand{\poi}{\mathbf{q_i}}
\newcommand{\tri}{\mathbf{t_i}}
\newcommand{\tbi}{\mathbf{t_b}}
\newcommand{\Tes}{\mathbf{T^{ee\to s}}}
\newcommand{\Twe}{\mathbf{T_i^{w \to ee}}}

\newcommand{\Gvertical}{\textbf{Vertical}}
\newcommand{\Ghorizontal}{\textbf{Horizontal}}
\newcommand{\Grollleftpitchup}{\textbf{Roll \& Pitch}}
\newcommand{\Gyawleftforward}{\textbf{Yaw}}

\newcommand{\Bfixed}{\textbf{FixedPose}}
\newcommand{\Beffonly}{\textbf{EfficiencyOnly}}
\newcommand{\Bcomfonly}{\textbf{ComfortOnly}}
\newcommand{\Bours}{\textbf{Comfort+Efficiency}}

\newcommand{\Mratings}{\textbf{Comfort}}
\newcommand{\Mease}{\textbf{Ease}}
\newcommand{\Mrank}{\textbf{Rank}}
\newcommand{\Msafety}{\textbf{Safety}}